\documentclass[letterpaper]{article}
\usepackage[preprint]{aaai2027}
\usepackage[hyphens]{url}
\usepackage{graphicx}
\usepackage{natbib}
\usepackage{caption}
\usepackage{algorithm}
\usepackage{algorithmic}
\usepackage{amsfonts}
\usepackage{comment}

\usepackage{newfloat}
\usepackage{listings}
\usepackage{makecell}

\DeclareCaptionStyle{ruled}{labelfont=normalfont,labelsep=colon,strut=off} 

\floatstyle{ruled}
\newfloat{listing}{tb}{lst}{}
\floatname{listing}{Listing}

\usepackage[table]{xcolor}
\definecolor{promptblue}{RGB}{238,246,255}
\definecolor{promptborder}{RGB}{120,160,210}

\usepackage{booktabs}
\usepackage{multirow}

\usepackage{amsmath}

\title{Look Before You Judge: Training-Free Region Mining for Grounded and Explainable Deepfake Detection}
\author {
    Chia-Ling Chen\textsuperscript{\rm 1}\equalcontrib,
    Yu-Ting Ta\textsuperscript{\rm 1}\equalcontrib,
    Jian-Yu Jiang-Lin\textsuperscript{\rm 1},
    Tai-Ming Huang\textsuperscript{\rm 1},
    Ling Lo\textsuperscript{\rm 2},
    Po-Ching Chen\textsuperscript{\rm 1},
    Yan-Tsung Wang\textsuperscript{\rm 1},
    Pei-Heng Li\textsuperscript{\rm 1},
    Ling Zou\textsuperscript{\rm 1},
    Hong-Han Shuai\textsuperscript{\rm 3},
    Wen-Huang Cheng\textsuperscript{\rm 1}\corresponding
}
\affiliations {
    \textsuperscript{\rm 1}National Taiwan University\\
    \textsuperscript{\rm 2}National Tsing Hua University\\
    \textsuperscript{\rm 3}National Yang Ming Chiao Tung University\\
    wenhuang@csie.ntu.edu.tw
}

\begin{document}
\maketitle
\begin{abstract}
Multimodal large language models (MLLMs) can explain deepfake verdicts in natural language, but such explanations are not necessarily visually grounded in the visual evidence underlying the prediction. A model may describe plausible artifacts inferred from language priors rather than from image evidence. Existing grounding methods improve visual reliance through decoding or attention interventions, but they generally strengthen grounding over the entire image, making them ill-suited for forensic artifacts that are subtle, spatially localized, and image-dependent. We propose Look Before You Judge, a training-free framework that formulates explainable deepfake detection as a sequential evidence acquisition process. Instead of directly predicting image authenticity from holistic visual reasoning, our framework first identifies image-specific candidate evidence regions by contrasting the MLLM's decoder-to-visual attention between an original image and its Gaussian-blurred counterpart. The identified regions are then inspected individually, and the resulting local evidence is integrated with the global image context before reaching a final verdict. The framework operates without manipulation masks, external forensic models, or parameter updates, making it directly applicable to off-the-shelf MLLMs. Across five open-source MLLMs on TriDF and MMTD-Set, our framework improves detection accuracy by up to 12.8\%, reduces CHAIR by up to 33.4\% and hallucination rate by up to 21.3\%, and outperforms representative training-free decoding and attention methods.

\end{abstract}
\begin{figure}[t]
    \centering
    \includegraphics[width=\columnwidth]{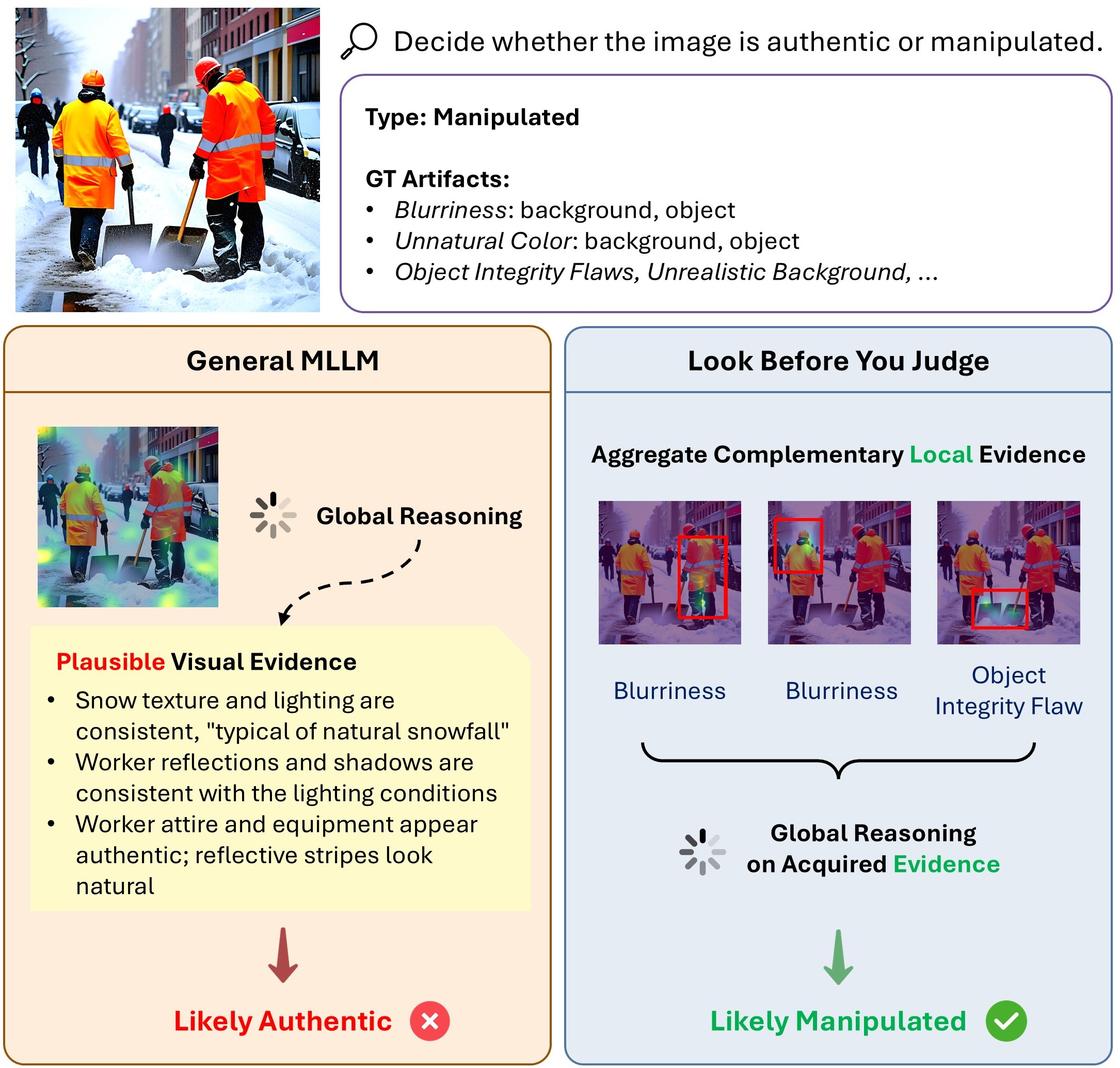}
    \caption{\textbf{Look Before You Judge.} A general MLLM produces an ungrounded explanation and an incorrect authentic verdict. Our framework instead inspects localized candidate regions and aggregates the resulting evidence to reach the correct manipulated verdict.}
    \label{fig:teaser}
\end{figure}

\section{Introduction}
\label{sec:intro}

Deep generative models have rapidly improved the realism of synthetic imagery, raising growing concerns about media integrity, identity verification, and digital forensics. Conventional deepfake detectors formulate deepfake detection as an image-level binary classification task, predicting whether an input is real or manipulated~\cite{chen2022self,dong2023implicit,haliassos2022leveraging,han2025towards,sun2026dfd}. However, a reliable forensic analysis requires more than a binary verdict. To support trustworthy decision making, a detector should provide concrete visual evidence that justifies its prediction. Motivated by this need, recent multimodal large language model (MLLM)-based forensic methods augment image-level prediction with natural-language explanations~\cite{jia2024chatgpt,Zou2025SurveyOA,Shi2025SHIELD}. Despite their impressive fluency, the explanations often fail to faithfully reflect the visual evidence underlying the prediction. As illustrated in Figure~\ref{fig:teaser}, a state-of-the-art MLLM may generate a plausible forensic explanation while overlooking annotated manipulation artifacts, ultimately producing an incorrect verdict. The key challenge of trustworthy deepfake detection is not generating explanations after prediction, but explicitly acquiring image-specific forensic evidence before making a forensic judgment.

Acquiring forensic evidence, however, is fundamentally more challenging than reasoning over semantic content. Unlike high-level semantic concepts, forensic evidence is typically weak, spatially localized, and expressed through subtle visual inconsistencies. Depending on the manipulation, the diagnostically useful cues may occur around facial structures in DeepFakes or around edited objects and regenerated regions in broader AIGC images. Under holistic image reasoning, these subtle forensic signals are easily overwhelmed by semantically dominant image content. Current MLLM-based forensic systems nevertheless attempt to localize evidence, perform reasoning, generate explanations, and predict authenticity within a unified inference process~\cite{xu2025fakeshield,huang2025sida,kang2025legion}, making evidence acquisition an implicit of reasoning rather than an explicit objective. Although recent grounding methods improve visual reliance through decoding or attention interventions~\cite{an2025mitigating,yin2025clearsight,tang2025seeing,zhuang2025vasparse,kim2025fuzzy}, they generally strengthen grounding over the entire image or assume informative regions are already available~\cite{wu2024controlmllm}. Consequently, existing approaches do not enable an off-the-shelf MLLM to autonomously identify image-specific forensic evidence before making a forensic judgment.

In this work, we formulate explainable deepfake detection as a sequential, spatially grounded evidence acquisition problem rather than a single-pass explanation task. Instead of directly predicting image authenticity from holistic visual reasoning, we advocate a \emph{look-before-you-judge} paradigm in which the model first identifies image-specific regions that are likely to contain diagnostically useful evidence, inspects these regions to collect manipulation-related observations, and finally integrates the resulting local evidence with the global image context before reaching a verdict. This formulation explicitly separates evidence acquisition from decision making, making local evidence inspection a prerequisite for reliable forensic reasoning rather than a by-product of explanation generation.

Following the formulation, we propose a training-free evidence acquisition framework for off-the-shelf MLLMs. Our key insight is that image regions containing forensic evidence exhibit distinctive attention responses to fine-grained visual perturbations. We exploit this property by contrasting the MLLM's decoder-to-visual attention between the original and blurred images, enabling the model to identify candidate evidence regions without additional supervision. The identified regions are subsequently inspected individually, and the resulting local observations are integrated with the global image context for final prediction. Since the entire pipeline relies solely on the deployed MLLM's test-time attention responses, our framework is self-contained and requires neither manipulation masks, predefined facial priors, external forensic encoders, learned localization modules, nor task-specific training.

Our contributions are summarized as follows:
\begin{itemize}
    \item We introduce a \emph{look-before-you-judge} formulation for explainable deepfake detection, which explicitly separates region discovery, local evidence inspection, and final image-level judgment. This formulation addresses the missing pre-decision inspection stage in existing MLLM-based forensic reasoning.

    \item Based on this formulation, we develop a fully training-free test-time evidence acquisition framework that identifies detail-sensitive forensic regions through blur-contrastive attention analysis and guides the MLLM to inspect them without manipulation masks, task-specific training, or learned localization modules.

    \item Extensive experiments across five open-source MLLM backbones and two benchmarks covering diverse deepfake manipulations and AIGC editing demonstrate consistent improvements in both detection accuracy and explanation grounding, validating the effectiveness and generalizability of the proposed framework.
\end{itemize}

\section{Related Work}
\subsection{From Detection to Local Evidence Inspection}
Early deepfake detectors formulate forgery detection as image-level binary classification, typically trained on large-scale datasets such as FaceForensics++~\cite{rossler2019faceforensics++} and DFDC~\cite{dolhansky2020deepfake}. Although effective in controlled settings, these models often rely on dataset-specific artifacts and may generalize poorly to unseen generators~\cite{rossler2019faceforensics++,dolhansky2020deepfake}. Later methods incorporate forgery-aware priors to improve robustness~\cite{liu2024forgeryaware, yang2025d3}, but still mainly provide image-level predictions without exposing local evidence.
Recent MLLM-based image forensic systems, such as FakeShield~\cite{xu2025fakeshield}, SIDA~\cite{huang2025sida}, and LEGION~\cite{kang2025legion}, extend detection toward interpretable forgery analysis by producing textual explanations and, in some cases, localization outputs. However, their localization and explanation are typically obtained through supervised training on annotated masks, artifact labels, or curated explanation data, and are generated jointly with the final decision. Our work neither learns localization from annotated data nor directly asks the MLLM for a global explanation. Instead, it first mines candidate forensic regions from the model's own test-time attention responses, enabling a \emph{look-before-you-judge} inspection of local evidence prior to the final judgment.

\begin{figure*}[!t]
    \centering
    \includegraphics[width=\textwidth]{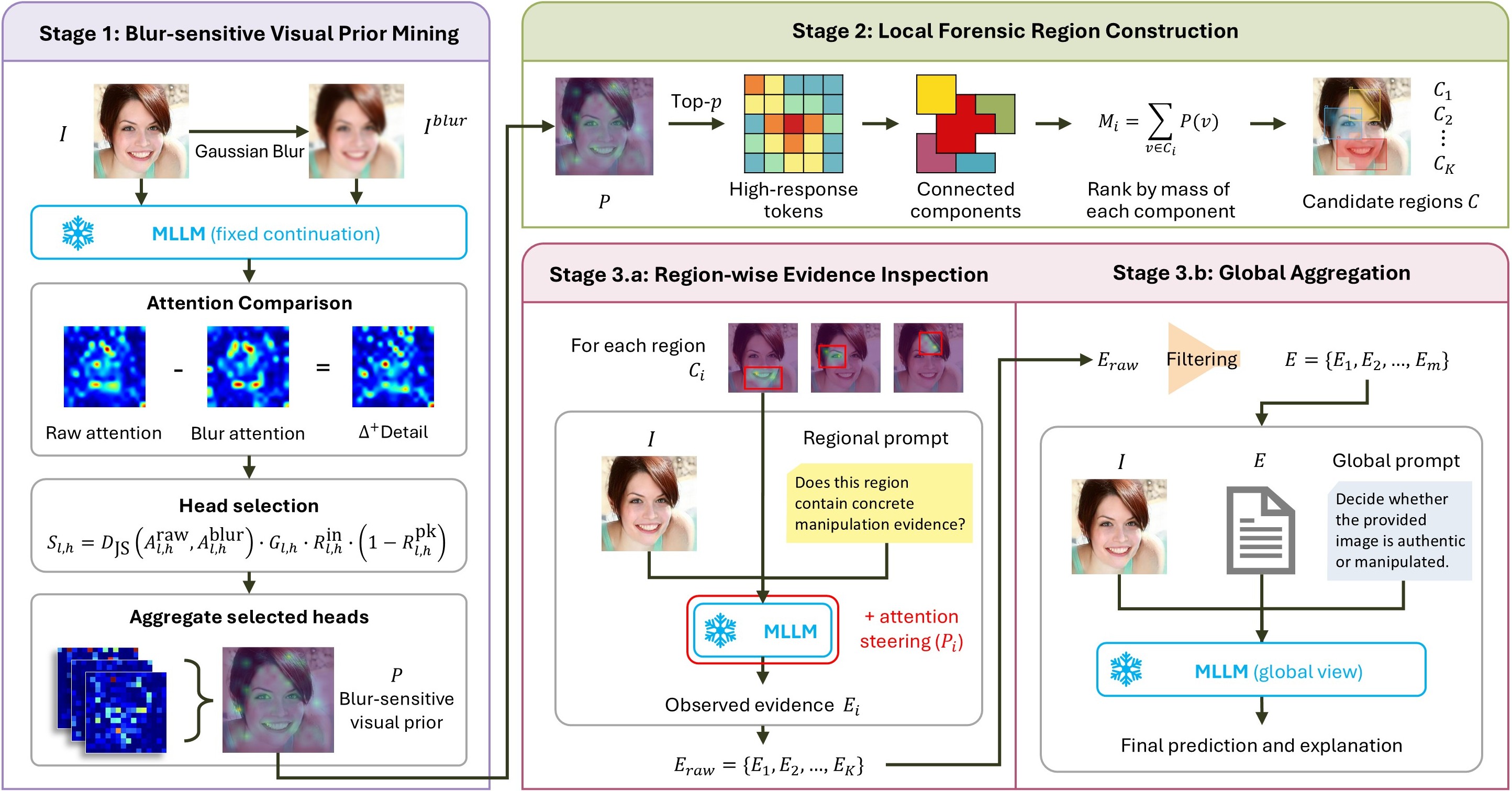}
    \caption{\textbf{Overview of \emph{Look Before You Judge}.} Stage 1 derives a blur-sensitive visual prior by contrasting output-to-visual attention between the input image and its blurred counterpart. Stage 2 constructs candidate forensic regions from high-response tokens. Stage 3 performs region-wise inspection with attention steering and aggregates the filtered evidence to produce the final prediction and explanation. The entire framework is training-free and requires only frozen MLLMs.}
    \label{fig:pipeline}
\end{figure*}

\subsection{Visual Grounding and Forensic Localization}

Recent studies have linked MLLM hallucination to insufficient visual grounding and an excessive reliance on language priors, motivating a growing body of training-free methods that intervene in attention or decoding at inference time~\cite{lu2026reallocating,basile2026head,tu2026attention,huang2024opera,leng2024mitigating}. While these approaches generally improve the model’s reliance on visual information, they predominantly operate at the whole-image level. Such global interventions may be suboptimal for forensic analysis, where discriminative artifacts are often sparse and confined to small regions of the image. ControlMLLM~\cite{wu2024controlmllm} enables attention steering toward designated image regions, but assumes that the regions of interest are specified in advance.

In parallel, existing forensic approaches extract localized evidence using frequency-domain priors~\cite{qian2020thinking}, gradient-based saliency~\cite{selvaraju2017grad}, or dedicated localization networks~\cite{li2020face,xu2025fakeshield,kang2025legion}. LAA-Net~\cite{nguyen2024laa} further reduces the dependence on explicit manipulation masks by learning attention over artifact-prone regions. Nevertheless, its localization behavior remains governed by task-specific training objectives and the artifact distributions represented in the training data.

These two lines of research therefore address complementary, yet incomplete, aspects of explainable forensic reasoning. Training-free grounding methods strengthen visual reliance without explicitly identifying where forensic evidence is located, whereas forensic localization methods typically depend on task-specific supervision or dedicated localization components. Our method bridges this gap by automatically deriving image-specific forensic regions from an MLLM’s inference-time attention responses. This enables region-wise evidence inspection without localization supervision, auxiliary localization modules, or parameter updates.

\section{Method}
We introduce a training-free framework that uses an MLLM's output-to-visual attention to determine where to inspect before making an image-level judgment. As shown in Figure~\ref{fig:pipeline}, it discovers and inspects detail-sensitive regions and verifies the evidence retained under an undirected global view.

The mined visual prior is \emph{authenticity-agnostic}: it serves only as a proposal signal for subsequent forensic inspection. Because all candidate regions and steering distributions are derived from test-time attention, the framework requires no manipulation masks, predefined facial regions, auxiliary modules, or parameter updates.

\subsection{Stage 1: Blur-Sensitive Visual Prior Mining}
Stage 1 mines a token-level \emph{blur-sensitive prior} that highlights visual tokens whose attention responses depend on fine-grained image detail. Because Gaussian blur suppresses such detail while largely preserving coarse image structure, reduced output-to-visual attention under blur provides a useful signal for forensic region mining.

We analyze decoder self-attention from generated response tokens to visual tokens, rather than the vision encoder's internal patch attention, as it reflects the visual evidence consulted while producing the forensic response.

Given a raw image $I$, we construct a Gaussian-blurred counterpart $I^{\mathrm{blur}}$. To avoid confounding visual changes with different textual continuations, we generate a response from $I$ and reuse its token sequence as a fixed, teacher-forced continuation for both inputs.

Let $l=1,\dots,L$ and $h=1,\dots,H$ index decoder layers and attention heads. For condition $c\in\{\mathrm{raw},\mathrm{blur}\}$, let $\alpha^{c, (l,h)}_{t,v}$ denote the post-softmax attention from response-token position $t$ to visual token $v$.

We average the output-to-visual attention over generated response positions:
\begin{equation}
    \bar{A}^{c,\mathrm{abs}}_{l,h}(v)
    =
    \frac{1}{|\mathcal{T}_{\mathrm{out}}|}
    \sum_{t\in\mathcal{T}_{\mathrm{out}}}
    \alpha^{c,(l,h)}_{t,v},
    \quad v=1,\dots,N_v,
\end{equation}
where $N_v$ denotes the number of visual tokens and $\mathcal{T}_{\mathrm{out}}$ the generated response-token positions. The superscript $\mathrm{abs}$ denotes the original post-softmax attention mass before renormalization over visual tokens.

From the aligned raw and blurred attention maps, we compute the positive raw--blur delta
\begin{equation}
    \Delta^+_{l,h}(v)
    =
    \mathrm{ReLU}
    \left(
    \bar{A}^{\mathrm{raw,abs}}_{l,h}(v)
    -
    \bar{A}^{\mathrm{blur,abs}}_{l,h}(v)
    \right),
\end{equation}

The reverse delta $\Delta^-_{l,h}$ is defined analogously by exchanging the raw and blurred terms. $\Delta^+_{l,h}$ provides the token-level signal used to construct the visual prior, while $\Delta^-_{l,h}$ is used only to assess the directionality of each head's attention shift.

\noindent\textbf{Head scoring.}
Not every head provides a reliable blur-sensitive response. We therefore score
heads by directional change, spatial redistribution, and spatial
reliability.

We first measure the fraction of directional change occurring in the desired $\mathrm{raw} > \mathrm{blur}$ direction:
\begin{equation}
    G_{l,h}
    =
    \frac{
    \sum_{v=1}^{N_v} \Delta^+_{l,h}(v)
    }{
    \sum_{v=1}^{N_v} \Delta^+_{l,h}(v)
    +
    \sum_{v=1}^{N_v} \Delta^-_{l,h}(v)
    +
    \epsilon
    },
\end{equation}
where $\epsilon > 0$ ensures numerical stability.

Let $\mathcal M_{\mathrm{in}}$ denote a valid-region mask excluding border tokens. For each head, $R^{\mathrm{in}}_{l,h}$ denotes the fraction of positive-delta mass within the valid region, while $R^{\mathrm{pk}}_{l,h}$ denotes the fraction concentrated on its highest-response token.

To measure spatial redistribution, we normalize each attention map over visual tokens:
\begin{equation}
    A^{c}_{l,h}(v)
    =
    \frac{
    \bar{A}^{c,\mathrm{abs}}_{l,h}(v)
    }{
    \sum_{u=1}^{N_v}
    \bar{A}^{c,\mathrm{abs}}_{l,h}(u)
    +
    \epsilon
    }.
\end{equation}
We combine these factors into the head score
\begin{equation}
    S_{l,h}
    =
    D_{\mathrm{JS}}
    \left(
    A^{\mathrm{raw}}_{l,h},
    A^{\mathrm{blur}}_{l,h}
    \right)
    \cdot
    G_{l,h}
    \cdot
    R^{\mathrm{in}}_{l,h}
    \cdot
    \left(
    1 - R^{\mathrm{pk}}_{l,h}
    \right),
\end{equation}
where $D_{\mathrm{JS}}$ denotes Jensen--Shannon divergence. This score favors heads with predominantly raw-over-blur change, meaningful spatial redistribution, and responses that are neither border-dominated nor concentrated on a single token.

We discard heads that fail fixed nondegeneracy checks and retain up to $K_H$
high-scoring heads under fixed diversity constraints, yielding the
image-specific set $\mathcal H$.

We normalize each selected positive-delta map to unit mass and average the resulting maps:
\begin{equation}
\bar{P}(v)
=
\frac{1}{|\mathcal{H}|}
\sum_{(l,h)\in \mathcal{H}}
\tilde{\Delta}^{+}_{l,h},
\ \ 
\tilde{\Delta}^{+}_{l,h}(v)
=
\frac{
\Delta^{+}_{l,h}(v)
}{
\sum_{u=1}^{N_v}
\Delta^{+}_{l,h}(u)
+
\epsilon
}.
\end{equation}
Per-head normalization prevents attention-scale differences from dominating
the pooled prior. After applying the valid-region mask, we renormalize $\bar P$ to obtain the Stage 1 prior $P\in\mathbb R^{N_v}$.

\subsection{Stage 2: Local Forensic Region Construction}
Stage 1 produces a token-level blur-sensitive prior $P$, whose high-response tokens may occupy multiple spatially separated locations. Stage 2 groups these responses into a small set of spatially coherent regions for independent inspection.

For each input, we map the one-dimensional prior onto a unified two-dimensional visual-token grid using the spatial layout provided by the corresponding backbone:
\begin{equation}
P \rightarrow P_{\mathrm{grid}} \in \mathbb{R}^{H_v \times W_v},
\quad N_v = H_v W_v,
\end{equation}
where $H_v$ and $W_v$ are the input-specific grid dimensions. This backbone-aware mapping preserves the spatial ordering of the visual tokens for subsequent region construction.

We retain the top-$p$ fraction of tokens, group spatially adjacent responses using connected-component analysis, discard degenerate components, and slightly expand the survivors for local context, yielding the candidate components $\{\tilde{C}_j\}$.

For each candidate component, we compute its total prior mass and the corresponding steering distribution:
\begin{equation}
\begin{aligned}
M_j &= \sum_{v \in \tilde{C}_j} P(v),
\qquad
P_j(v) &=
\begin{cases}
\dfrac{P(v)}{M_j}, & v \in \tilde{C}_j,\\[4pt]
0, & v \notin \tilde{C}_j.
\end{cases}
\end{aligned}
\end{equation}

Here $M_j$ measures the component's contribution to the image-level blur-sensitive prior and is used for region ranking, while $P_j$ preserves its relative prior weights and satisfies $\sum_vP_j(v)=1$ for Stage 3 steering.

We rank the candidate components by $M_j$ and retain up to $K$ regions:
\begin{equation}
\mathcal{C}
=
\operatorname{Top\text{-}}\!K_{j}\!\left(M_j\right)
=
\{C_1, C_2, \dots, C_{K'}\},
\quad
K' \leq K,
\end{equation}

where $K'<K$ when fewer than $K$ valid components remain. After ranking, the retained components and their steering distributions are reindexed as $\{(C_i,P_i)\}_{i=1}^{K'}$ for region-wise inspection in Stage 3.

\begin{table*}[t]
\centering
\small
\begin{tabular}{l ccccc cccc}
\toprule
\multirow{3}{*}{\textbf{Method}}
& \multicolumn{5}{c}{\textbf{TriDF}}
& \multicolumn{4}{c}{\textbf{MMTD-Set}} \\
\cmidrule(lr){2-6}
\cmidrule(lr){7-10}
& \multirow{2}{*}{\textit{ACC} $\uparrow$}
& \multirow{2}{*}{\textit{Cover} $\uparrow$}
& \multirow{2}{*}{\textit{CHAIR} $\downarrow$}
& \multirow{2}{*}{\textit{Hal} $\downarrow$}
& \multirow{2}{*}{$F^{0.5}$ $\uparrow$}
& \multicolumn{2}{c}{\textit{DeepFake}}
& \multicolumn{2}{c}{\textit{AIGC-Editing}} \\
\cmidrule(lr){7-8}
\cmidrule(lr){9-10}
& & & & &
& \textit{ACC} $\uparrow$
& \textit{F1} $\uparrow$
& \textit{ACC} $\uparrow$
& \textit{F1} $\uparrow$ \\
\midrule
\multicolumn{10}{c}{\textit{\textbf{Open-source models}}} \\
\midrule
InternVL-3.5-8B
& 0.4176 & 0.0270 & 0.9745 & 1.0000 & 0.0296 & 0.5235 & 0.4936 & 0.5200 & 0.4419 \\
\rowcolor{gray!20}
InternVL-3.5-8B + Ours
& \textbf{0.5458} & \textbf{0.2239} & \textbf{0.6407} & \textbf{0.7875} & \textbf{0.2564} & \textbf{0.5591} & \textbf{0.6278} & \textbf{0.5260} & \textbf{0.5917} \\
\midrule
InternVL-3.5-14B
& 0.4213 & 0.0395 & 0.9763 & 0.9980 & 0.0217 & 0.5300 & 0.5488 & 0.5105 & 0.4872 \\
\rowcolor{gray!20}
InternVL-3.5-14B + Ours
& \textbf{0.5324} & \textbf{0.2045} & \textbf{0.7963} & \textbf{0.9449} & \textbf{0.1587} & \textbf{0.5776} & \textbf{0.5496} & \textbf{0.5525} & \textbf{0.6193} \\
\midrule
MiMo-VL-7B
& 0.5650 & \textbf{0.2280} & 0.6539 & 0.8739 & \textbf{0.2914} & 0.6046 & 0.5423 & \textbf{0.5935} & 0.4389 \\
\rowcolor{gray!20}
MiMo-VL-7B + Ours
& \textbf{0.6086} & 0.1110 & \textbf{0.6536} & \textbf{0.7271} & 0.2144 & \textbf{0.6326} & \textbf{0.6011} & 0.5655 & \textbf{0.4749} \\
\midrule
Qwen3-VL-8B-Instruct
& 0.6207 & \textbf{0.2557} & 0.8073 & 0.9993 & \textbf{0.2022} & 0.6141 & 0.5952 & 0.4940 & 0.4960 \\
\rowcolor{gray!20}
Qwen3-VL-8B-Instruct + Ours
& \textbf{0.6408} & 0.1875 & \textbf{0.7566} & \textbf{0.9349} & 0.2020 & \textbf{0.6181} & \textbf{0.6294} & \textbf{0.5725} & \textbf{0.5479} \\
\midrule
Qwen3.5-9B
& 0.6969 & 0.0700 & 0.9381 & 0.9889 & 0.0552 & 0.5195 & 0.6524 & 0.4885 & 0.6207 \\
\rowcolor{gray!20}
Qwen3.5-9B + Ours
& \textbf{0.6972} & \textbf{0.1088} & \textbf{0.8722} & \textbf{0.9564} & \textbf{0.1108} & \textbf{0.6066} & \textbf{0.6851} & \textbf{0.6130} & \textbf{0.6387} \\
\midrule
\multicolumn{10}{c}{\textit{\textbf{Commercial models}}} \\
\midrule
GPT-5
& 0.6573 & 0.2714 & 0.6982 & 0.9651 & 0.2919
& 0.7727 & 0.7471 & 0.5786 & 0.2936 \\
Gemini 2.5-Pro
& 0.7311 & 0.4208 & 0.5571 & 0.9332 & 0.4258
& 0.5998 & 0.6664 & 0.7063 & 0.7302 \\
Claude Sonnet 4.5
& 0.6240 & 0.3988 & 0.7235 & 0.9980 & 0.2908
& 0.5509 & 0.6513 & 0.5804 & 0.5508 \\
\bottomrule
\end{tabular}
\caption{\textbf{Overall Quantitative Comparison.} Results across five open-source MLLMs on TriDF and MMTD-Set, with commercial MLLMs reported as performance references.}
\label{tab:tridf_mmtd}
\end{table*}

\subsection{Stage 3: Region-Wise Evidence Inspection and Global Aggregation}
Stage 3 converts the retained regions into structured forensic evidence and aggregates the filtered evidence into an image-level verdict. It comprises region-wise inspection (Stage 3.a), then evidence filtering and global aggregation (Stage 3.b).

\subsubsection{Stage 3.a: Region-Wise Evidence Inspection}
For each candidate region $C_i \in \mathcal{C}$, we steer generation using its distribution $P_i$ from Stage 2. Following AttnReal~\cite{tu2026attention}, we recycle attention from generated-response positions, but redistribute it according to $P_i$ rather than uniformly over all visual tokens.

Let $\mathcal H^\star$ denote all attention heads within a model-specific decoder-layer range $\mathcal L^\star$. This intervention set is distinct from the mined set $\mathcal H$, used only to construct the Stage 1 prior.

Following the sink-selection rule of AttnReal, let
$\mathcal S_t^{(l,h)}$ denote positions selected from the
generated-response history. With retention coefficient $\rho\in[0,1)$, the recycled mass is
\begin{equation}
    m^{(l,h)}_{t}
    =
    (1-\rho)
    \sum_{k \in \mathcal{S}^{(l,h)}_t}
    \alpha^{(l,h)}_{t,k}.
\end{equation}

Let $\mathcal V$ denote the visual-token key positions, with $P_i(k)$ denoting the weight of the visual token at position $k\in\mathcal V$. For $(l,h)\in\mathcal{H}^\star$, we update
\begin{equation}
\tilde{\alpha}^{(l,h)}_{t,k}
=
\begin{cases}
\rho\,\alpha^{(l,h)}_{t,k}, & k \in \mathcal{S}^{(l,h)}_t, \\
\alpha^{(l,h)}_{t,k} + m^{(l,h)}_{t}\, P_i(k), & k \in \mathcal{V}, \\
\alpha^{(l,h)}_{t,k}, & \text{otherwise}.
\end{cases}
\end{equation}

Because the sink and visual positions are disjoint and $\sum_vP_i(v)=1$, the removed and added masses are equal, preserving the attention-row sum while redistributing attention toward $C_i$.

The same region-agnostic prompt $q_{\mathrm{loc}}$ is used for every region.
It asks the model to report a local irregularity and assess its evidential
strength without issuing an image-level verdict. The structured response is

\begin{equation}
    E_i
    =
    (o_i, e_i, \gamma_i, a_i, n_i),
\end{equation}
consisting of a localized observation, manipulation-evidence label, confidence, artifact category, and optional natural explanation. The responses form $\mathcal E_{\mathrm{raw}}=\{E_i\}_{i=1}^{K'}$.

\subsubsection{Stage 3.b: Evidence Filtering and Global Aggregation}
\noindent\textbf{Evidence Filtering.}
Region-wise inspection may produce unreliable or redundant observations. We process $\mathcal{E}_{\mathrm{raw}}$ in descending region-prior order, filtering such responses and retaining the higher-priority observation when duplicates occur. The remaining observations form $\mathcal{E}_{\mathrm{cand}}$, with their contribution calibrated according to evidential strength before global aggregation.

\noindent\textbf{Final Aggregation.}
We construct an aggregation prompt $q_{\mathrm{agg}}(\mathcal{E}_{\mathrm{cand}})$ from the authenticity instruction and filtered observations, presenting them as tentative cues to be verified against the full image. The final prediction is
\begin{equation}
    \hat{y} = g\big(I,\, q_{\mathrm{agg}}(\mathcal{E}_{\mathrm{cand}})\big),
\end{equation}
where $g$ denotes standard inference with the same model but without attention steering, thereby restoring the global context needed to verify and integrate the local observations.

\section{Experiments}

\subsection{Experimental Setup}
\noindent\textbf{Datasets and Evaluation Metrics.}
We evaluate our framework on TriDF Type-B \textless OEQ\textgreater{}~\cite{jiang2026tridf} and MMTD-Set~\cite{xu2025fakeshield}. TriDF provides artifact-level annotations, enabling joint evaluation of authenticity prediction and explanation grounding. We report accuracy (ACC), artifact coverage (Cover), CHAIR, hallucination rate (Hal), and $F^{0.5}$ on TriDF. Cover measures the recall of annotated artifacts, whereas CHAIR and Hal quantify unsupported artifact claims. The $F^{0.5}$ score summarizes the trade-off between artifact precision and coverage, with greater emphasis on precision. On MMTD-Set, we report ACC and F1 on the DeepFake and AIGC-Editing subsets.

\noindent\textbf{Backbone Models.}
We evaluate five off-the-shelf MLLMs: InternVL-3.5-8B, InternVL-3.5-14B~\cite{wang2025internvl3}, Qwen3-VL-8B-Instruct~\cite{bai2025qwen3vl}, Qwen3.5-9B~\cite{qwen3_5}, and MiMo-VL-7B~\cite{mimovl}. For each backbone, we compare vanilla inference with the proposed test-time evidence acquisition framework.


\noindent\textbf{Implementation Details.}
All experiments are conducted in a fully training-free setting. The same three-stage inference procedure is applied across all MLLM backbones. Additional implementation details and hyperparameter settings are provided in the supplementary material.

\begin{table}[t]
\centering
\small
\setlength{\tabcolsep}{2pt}

\resizebox{0.9\columnwidth}{!}{%
\begin{tabular}{lccccc}
\toprule
\multirow{2}{*}{\textbf{Method}}
& \multicolumn{5}{c}{\textbf{TriDF}} \\
\cmidrule(lr){2-6}
& \textit{ACC} $\uparrow$
& \textit{Cover} $\uparrow$
& \textit{CHAIR} $\downarrow$
& \textit{Hal} $\downarrow$
& $F^{0.5}$ $\uparrow$ \\
\midrule
Vanilla
& 0.4176 & 0.0270 & 0.9745 & 1.0000 & 0.0296 \\
\quad + VCD
& 0.4241 & 0.0372 & 0.9777 & 1.0000 & 0.0212 \\
\quad + AttnReal
& 0.4206 & 0.0719 & 0.9750 & 0.9993 & 0.0246 \\
\rowcolor{gray!20}
\quad + Ours
& \textbf{0.5458}
& \textbf{0.2239}
& \textbf{0.6407}
& \textbf{0.7875}
& \textbf{0.2564} \\
\bottomrule
\end{tabular}
}
\caption{\textbf{Comparison with Training-Free Grounding.} Results on TriDF using InternVL-3.5-8B.}
\label{tab:hallucination_reduction}
\end{table}

\subsection{Overall Quantitative Comparison}
We first evaluate the generalizability of the proposed training-free evidence acquisition framework across different MLLM backbones, benchmark datasets, and model scales. Table~\ref{tab:tridf_mmtd} summarizes the overall quantitative results.

\noindent\textbf{Generalization across MLLM Backbones.}
Our framework consistently improves image-level accuracy while substantially reducing CHAIR and hallucination rates across all five evaluated open-source MLLM backbones, demonstrating robust generalization across diverse architectures. For example, InternVL-3.5-8B improves accuracy from 0.418 to 0.546 while reducing CHAIR from 0.975 to 0.641. Although artifact coverage decreases for some models, these reductions are consistently accompanied by lower CHAIR and hallucination rates, indicating a shift toward more selective and better-supported artifact descriptions rather than exhaustive but potentially unsupported enumeration.

\noindent\textbf{Generalization across Benchmarks.}
We further evaluate our framework on MMTD-Set. Our framework improves both ACC and F1 across nearly all model--subset combinations on the DeepFake and AIGC-Editing subsets. Together with the results on TriDF, which also covers diverse manipulation types, these consistent gains demonstrate that the framework generalizes across benchmark datasets and heterogeneous manipulation scenarios.

\noindent\textbf{Comparison with Commercial MLLMs.}
For reference, we also compare our framework against several commercial MLLMs evaluated under vanilla inference. While these proprietary models generally outperform open-source MLLMs under vanilla inference, applying our framework substantially narrows the gap. In several cases, the enhanced open-source models achieve comparable or even superior explanation grounding despite requiring neither additional training nor proprietary components. These results suggest that effective test-time evidence acquisition can be as critical as model scaling for explainable forensic reasoning.

\subsection{Comparison with Training-Free Visual Grounding}

We further compare our framework with VCD~\cite{leng2024mitigating} and AttnReal~\cite{tu2026attention}, two representative training-free visual grounding methods that improve visual reliance during inference without additional training. As shown in Table~\ref{tab:hallucination_reduction}, all methods are evaluated using the same InternVL-3.5-8B backbone and TriDF setting. VCD and AttnReal yield only marginal gains over vanilla inference: improvements in ACC and Cover are limited, CHAIR and Hal remain nearly unchanged, and $F^{0.5}$ decreases. In contrast, our framework improves all five metrics, reducing CHAIR from $0.9745$ to $0.6407$ together with a marked decrease in Hal. These results indicate that globally strengthening visual reliance is insufficient for deepfake forensics; effective grounding instead requires spatially selective region discovery and inspection before the final judgment.

\subsection{Ablation Study}
\label{sec:ablation}
\begin{table}[t]
\centering
\small
\setlength{\tabcolsep}{2pt}
\resizebox{\columnwidth}{!}{%
\begin{tabular}{lccccc}
\toprule
\multirow{2}{*}{\textbf{Method}}
& \multicolumn{5}{c}{\textbf{TriDF}} \\
\cmidrule(lr){2-6}
& \textit{ACC} $\uparrow$
& \textit{Cover} $\uparrow$
& \textit{CHAIR} $\downarrow$
& \textit{Hal} $\downarrow$
& $F^{0.5}$ $\uparrow$ \\
\midrule
Vanilla
& 0.4176 & 0.0270 & 0.9745 & 1.0000 & 0.0296 \\

Prompt-only
& 0.5113 & 0.1369 & 0.8545 & 0.9927 & 0.1340 \\

w/o blur contrast
& 0.4867 & 0.1920 & 0.8600 & 0.9568 & 0.1132 \\

w/o component steering
& 0.4884 & 0.1890 & 0.8574 & 0.9548 & 0.1182 \\

\rowcolor{gray!20}
Ours
& \textbf{0.5458}
& \textbf{0.2239}
& \textbf{0.6407}
& \textbf{0.7875}
& \textbf{0.2564} \\
\bottomrule
\end{tabular}
}
\caption{\textbf{Ablation Study.} Effects of evidence-before-judgment prompting, blur-contrastive region mining, and component-wise steering on TriDF.}
\label{tab:ablation_tridf}
\end{table}

\begin{figure}[t]
    \centering
    \includegraphics[width=\columnwidth]{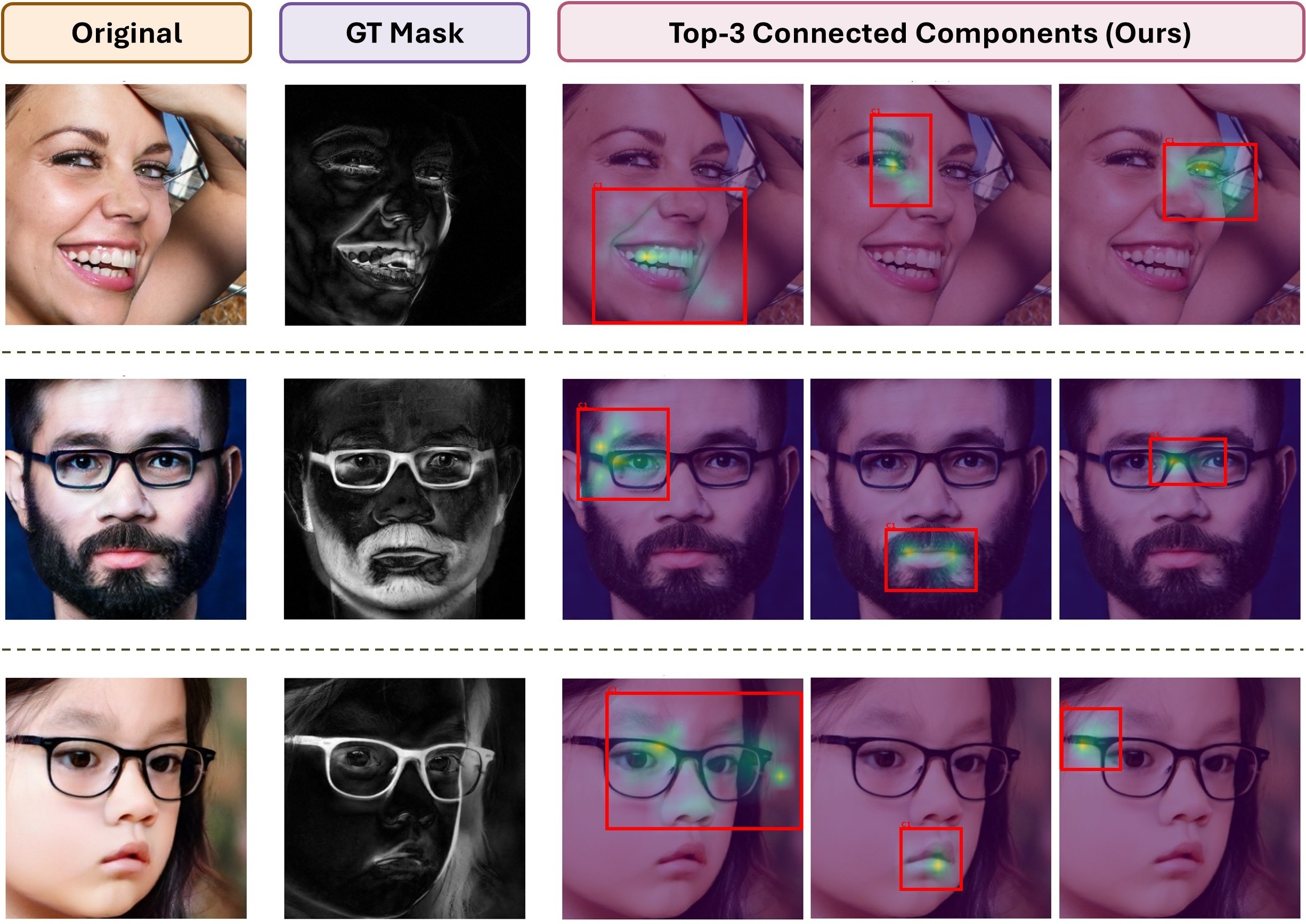}
    \caption{\textbf{Qualitative Spatial Grounding.} Candidate regions discovered by Stages~1 and 2 align with annotated tampered areas without localization supervision.}
    \label{fig:spatial_grounding}
\end{figure}

\begin{figure*}[t]
    \centering
    \includegraphics[width=\textwidth]
    {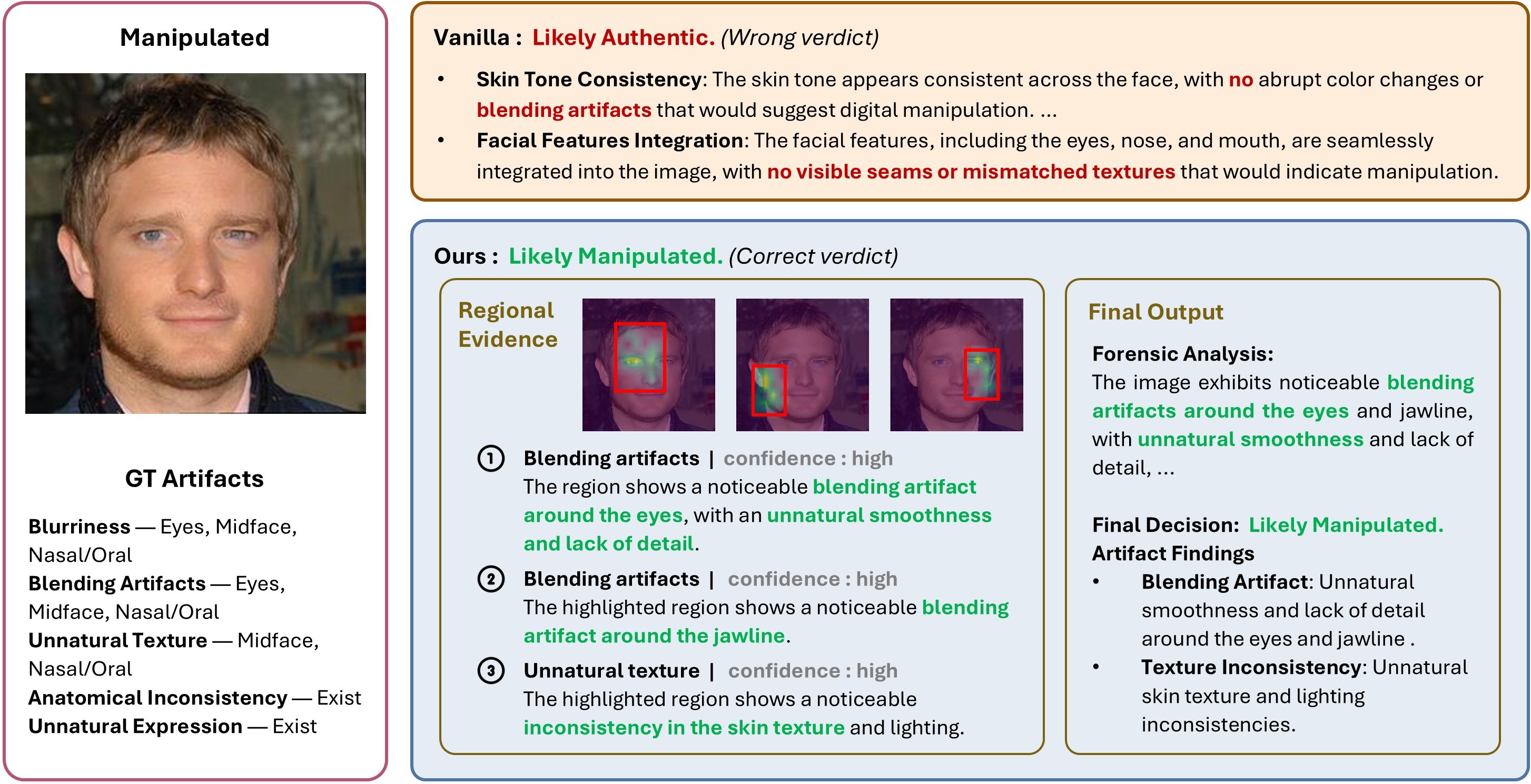}
    \caption{\textbf{From Global Error to Local Evidence.} \textbf{Top:} Vanilla inference overlooks the annotated blending and texture artifacts and incorrectly predicts the image as authentic. \textbf{Bottom:} Our framework discovers three candidate regions, inspects them independently, and aggregates the retained local evidence under a global view to produce the correct manipulated verdict.}
    \label{fig:Qualitative}
\end{figure*}

We conduct an ablation study on TriDF using InternVL-3.5-8B to evaluate the \emph{look-before-you-judge} procedure, blur-contrastive region mining, and connected-component steering. Table~\ref{tab:ablation_tridf} compares \emph{Vanilla}, \emph{Prompt-only}, \emph{w/o blur contrast}, \emph{w/o component steering}, and the full framework.

\noindent\textbf{Prompted evidence is not necessarily grounded.}
The prompt-only variant improves detection and artifact coverage over \emph{Vanilla}, but leaves hallucination metrics largely unchanged. This shows that eliciting more evidence does not ensure that the reported observations are visually supported.

\noindent\textbf{Region discovery matters.}
Using raw attention instead of the blur-contrastive prior underperforms even the prompt-only variant, suggesting that globally salient responses do not reliably expose subtle forensic cues. Effective inspection therefore depends on first discovering informative, detail-sensitive regions.

\noindent\textbf{Separate inspection improves grounding.}
Without component-level steering, a single global prior mixes spatially distinct cues and weakens focused inspection. In contrast, the full framework achieves the best overall performance, validating the complete \emph{look-before-you-judge} pipeline of region discovery, separate inspection, and evidence aggregation.

\subsection{Qualitative Spatial Grounding}

We qualitatively examine whether regions discovered before judgment correspond to actual manipulation areas. Figure~\ref{fig:spatial_grounding} visualizes the top candidate regions produced by blur-contrastive region mining and connected-component grouping on fake images from the DeepFake subset of MMTD-Set. Although the framework is not trained for localization and does not access ground-truth masks during inference, the discovered regions frequently cover manipulated facial boundaries and locally edited structures. These examples suggest that the proposed procedure produces meaningful inspection targets for acquiring local evidence before final judgment.

\subsection{Qualitative Analysis}

Figure~\ref{fig:Qualitative} shows how the proposed \emph{look-before-you-judge} procedure corrects an erroneous global judgment. Vanilla inference predicts the manipulated image as authentic based on broad claims of consistent skin tone and seamless facial integration. In contrast, our framework identifies candidate regions around the eyes, facial boundary, and jawline, where region-wise inspection reveals blending artifacts, abnormal smoothness, and texture inconsistencies. Aggregating these locally grounded observations under the global view leads to the correct manipulated verdict. Additional examples and failure cases are provided in the supplementary material.

\section{Conclusion}
We introduce \emph{Look Before You Judge}, a formulation that casts explainable deepfake detection as a sequential evidence acquisition problem rather than a single-pass prediction task. Based on this formulation, we develop a fully training-free test-time evidence acquisition framework that identifies detail-sensitive candidate regions from blur-contrastive attention responses, inspects them individually, and integrates the resulting local evidence into a final image-level judgment. The framework operates entirely at inference time on frozen MLLMs, requiring neither localization supervision, auxiliary modules, nor parameter updates.

Experiments across five open-source MLLMs and two benchmarks demonstrate consistent improvements in detection accuracy and explanation grounding. Together with the ablation and qualitative analyses, these results support the central premise of \emph{Look Before You Judge}: trustworthy forensic reasoning should first identify where potential evidence resides, then examine it locally, and only afterward reach a global conclusion. We hope this work motivates future research on evidence-centric DeepFake detection, where explicit evidence acquisition becomes a fundamental component of trustworthy and interpretable decision making beyond deepfake detection.

\bibliography{references}

@String(CVPR  = {IEEE Conf. Comput. Vis. Pattern Recog.})

@inproceedings{xu2025fakeshield,
  title={Fakeshield: Explainable image forgery detection and localization via multi-modal large language models},
  author={Xu, Zhipei and Zhang, Xuanyu and Li, Runyi and Tang, Zecheng and Huang, Qing and Zhang, Jian},
  booktitle={International Conference on Learning Representations},
  volume={2025},
  pages={31186--31216},
  year={2025}
}

@inproceedings{huang2025sida,
  title={Sida: Social media image deepfake detection, localization and explanation with large multimodal model},
  author={Huang, Zhenglin and Hu, Jinwei and Li, Xiangtai and He, Yiwei and Zhao, Xingyu and Peng, Bei and Wu, Baoyuan and Huang, Xiaowei and Cheng, Guangliang},
  booktitle={Proceedings of the Computer Vision and Pattern Recognition Conference},
  pages={28831--28841},
  year={2025}
}

@inproceedings{lu2026reallocating,
  title={Reallocating Attention Across Layers to Reduce Multimodal Hallucination},
  author={Lu, Haolang and Chu, Bolun and Fu, WeiYe and Nan, Guoshun and Liu, Junning and Pan, Minghui and Li, Qiankun and Yu, Yi and Wang, Hua and Wang, Kun},
  booktitle={Proceedings of the IEEE/CVF Conference on Computer Vision and Pattern Recognition},
  pages={4157--4167},
  year={2026}
}

@article{tu2026attention,
  title={Attention reallocation: Towards zero-cost and controllable hallucination mitigation of mllms},
  author={Tu, Chongjun and Ye, Peng and Zhou, Dongzhan and Bai, Lei and Yu, Gang and Chen, Tao and Ouyang, Wanli},
  journal={International Journal of Computer Vision},
  volume={134},
  number={1},
  pages={22},
  year={2026},
  publisher={Springer}
}

@inproceedings{jiang2026tridf,
  title={TriDF: Evaluating Perception, Detection, and Hallucination for Interpretable DeepFake Detection},
  author={Jiang-Lin, Jian-Yu and Huang, Kang-Yang and Zou, Ling and Lo, Ling and Yang, Sheng-Ping and Tseng, Yu-Wen and Lin, Kun-Hsiang and Chen, Chia-Ling and Ta, Yu-Ting and Wang, Yan-Tsung and others},
  booktitle={Proceedings of the IEEE/CVF Conference on Computer Vision and Pattern Recognition},
  pages={17087--17098},
  year={2026}
}

@inproceedings{rossler2019faceforensics++,
  title={Faceforensics++: Learning to detect manipulated facial images},
  author={Rossler, Andreas and Cozzolino, Davide and Verdoliva, Luisa and Riess, Christian and Thies, Justus and Nie{\ss}ner, Matthias},
  booktitle={Proceedings of the IEEE/CVF international conference on computer vision},
  pages={1--11},
  year={2019}
}

@article{dolhansky2020deepfake,
  title={The deepfake detection challenge (dfdc) dataset},
  author={Dolhansky, Brian and Bitton, Joanna and Pflaum, Ben and Lu, Jikuo and Howes, Russ and Wang, Menglin and Ferrer, Cristian Canton},
  journal={arXiv preprint arXiv:2006.07397},
  year={2020}
}

@inproceedings{kang2025legion,
  title={Legion: Learning to ground and explain for synthetic image detection},
  author={Kang, Hengrui and Wen, Siwei and Wen, Zichen and Ye, Junyan and Li, Weijia and Feng, Peilin and Zhou, Baichuan and Wang, Bin and Lin, Dahua and Zhang, Linfeng and others},
  booktitle={Proceedings of the IEEE/CVF International Conference on Computer Vision},
  pages={18937--18947},
  year={2025}
}

@inproceedings{nguyen2024laa,
  title={Laa-net: Localized artifact attention network for quality-agnostic and generalizable deepfake detection},
  author={Nguyen, Dat and Mejri, Nesryne and Singh, Inder Pal and Kuleshova, Polina and Astrid, Marcella and Kacem, Anis and Ghorbel, Enjie and Aouada, Djamila},
  booktitle={Proceedings of the IEEE/CVF Conference on Computer Vision and Pattern Recognition},
  pages={17395--17405},
  year={2024}
}

@inproceedings{leng2024mitigating,
  title={Mitigating object hallucinations in large vision-language models through visual contrastive decoding},
  author={Leng, Sicong and Zhang, Hang and Chen, Guanzheng and Li, Xin and Lu, Shijian and Miao, Chunyan and Bing, Lidong},
  booktitle={Proceedings of the IEEE/CVF Conference on Computer Vision and Pattern Recognition},
  pages={13872--13882},
  year={2024}
}

@inproceedings{huang2024opera,
  title={Opera: Alleviating hallucination in multi-modal large language models via over-trust penalty and retrospection-allocation},
  author={Huang, Qidong and Dong, Xiaoyi and Zhang, Pan and Wang, Bin and He, Conghui and Wang, Jiaqi and Lin, Dahua and Zhang, Weiming and Yu, Nenghai},
  booktitle={Proceedings of the IEEE/CVF Conference on Computer Vision and Pattern Recognition},
  pages={13418--13427},
  year={2024}
}

@article{wu2024controlmllm,
  title={Controlmllm: Training-free visual prompt learning for multimodal large language models},
  author={Wu, Mingrui and Cai, Xinyue and Ji, Jiayi and Li, Jiale and Huang, Oucheng and Fei, Hao and Jiang, Guannan and Sun, Xiaoshuai and Ji, Rongrong},
  journal={Advances in Neural Information Processing Systems},
  volume={37},
  pages={45206--45234},
  year={2024}
}

@article{basile2026head,
  title={Head pursuit: Probing attention specialization in multimodal transformers},
  author={Basile, Lorenzo and Maiorca, Valentino and Doimo, Diego and Locatello, Francesco and Cazzaniga, Alberto},
  journal={Advances in Neural Information Processing Systems},
  volume={38},
  pages={112351--112383},
  year={2026}
}

@inproceedings{qian2020thinking,
  title={Thinking in frequency: Face forgery detection by mining frequency-aware clues},
  author={Qian, Yuyang and Yin, Guojun and Sheng, Lu and Chen, Zixuan and Shao, Jing},
  booktitle={European conference on computer vision},
  pages={86--103},
  year={2020},
  organization={Springer}
}

@inproceedings{selvaraju2017grad,
  title={Grad-cam: Visual explanations from deep networks via gradient-based localization},
  author={Selvaraju, Ramprasaath R and Cogswell, Michael and Das, Abhishek and Vedantam, Ramakrishna and Parikh, Devi and Batra, Dhruv},
  booktitle={Proceedings of the IEEE international conference on computer vision},
  pages={618--626},
  year={2017}
}

@inproceedings{li2020face,
  title={Face x-ray for more general face forgery detection},
  author={Li, Lingzhi and Bao, Jianmin and Zhang, Ting and Yang, Hao and Chen, Dong and Wen, Fang and Guo, Baining},
  booktitle={Proceedings of the IEEE/CVF conference on computer vision and pattern recognition},
  pages={5001--5010},
  year={2020}
}

@inproceedings{han2025towards,
  title={Towards more general video-based deepfake detection through facial component guided adaptation for foundation model},
  author={Han, Yue-Hua and Huang, Tai-Ming and Hua, Kai-Lung and Chen, Jun-Cheng},
  booktitle={Proceedings of the IEEE/CVF conference on computer vision and pattern recognition},
  pages={22995--23005},
  year={2025}
}

@inproceedings{sun2026dfd,
  title={DFD-HR: Generalizable Deepfake Detection via Hierarchical Routing Learning},
  author={Sun, Jiamu and Yan, Zhiyuan and Zhang, Ke-Yue and Yao, Taiping and Ding, Shouhong},
  booktitle={Proceedings of the IEEE/CVF Conference on Computer Vision and Pattern Recognition},
  pages={13984--13995},
  year={2026}
}

@inproceedings{chen2022self,
  title={Self-supervised learning of adversarial example: Towards good generalizations for deepfake detection},
  author={Chen, Liang and Zhang, Yong and Song, Yibing and Liu, Lingqiao and Wang, Jue},
  booktitle={Proceedings of the IEEE/CVF conference on computer vision and pattern recognition},
  pages={18710--18719},
  year={2022}
}

@inproceedings{dong2023implicit,
  title={Implicit identity leakage: The stumbling block to improving deepfake detection generalization},
  author={Dong, Shichao and Wang, Jin and Ji, Renhe and Liang, Jiajun and Fan, Haoqiang and Ge, Zheng},
  booktitle={Proceedings of the IEEE/CVF conference on computer vision and pattern recognition},
  pages={3994--4004},
  year={2023}
}

@inproceedings{haliassos2022leveraging,
  title={Leveraging real talking faces via self-supervision for robust forgery detection},
  author={Haliassos, Alexandros and Mira, Rodrigo and Petridis, Stavros and Pantic, Maja},
  booktitle={Proceedings of the IEEE/CVF conference on computer vision and pattern recognition},
  pages={14950--14962},
  year={2022}
}

@inproceedings{an2025mitigating,
  title={Mitigating object hallucinations in large vision-language models with assembly of global and local attention},
  author={An, Wenbin and Tian, Feng and Leng, Sicong and Nie, Jiahao and Lin, Haonan and Wang, QianYing and Chen, Ping and Zhang, Xiaoqin and Lu, Shijian},
  booktitle={Proceedings of the Computer Vision and Pattern Recognition Conference},
  pages={29915--29926},
  year={2025}
}

@inproceedings{yin2025clearsight,
  title={Clearsight: Visual signal enhancement for object hallucination mitigation in multimodal large language models},
  author={Yin, Hao and Si, Guangzong and Wang, Zilei},
  booktitle={Proceedings of the Computer Vision and Pattern Recognition Conference},
  pages={14625--14634},
  year={2025}
}

@inproceedings{tang2025seeing,
  title={Seeing far and clearly: Mitigating hallucinations in mllms with attention causal decoding},
  author={Tang, Feilong and Liu, Chengzhi and Xu, Zhongxing and Hu, Ming and Huang, Zile and Xue, Haochen and Chen, Ziyang and Peng, Zelin and Yang, Zhiwei and Zhou, Sijin and others},
  booktitle={Proceedings of the Computer Vision and Pattern Recognition Conference},
  pages={26147--26159},
  year={2025}
}

@inproceedings{zhuang2025vasparse,
  title={Vasparse: Towards efficient visual hallucination mitigation via visual-aware token sparsification},
  author={Zhuang, Xianwei and Zhu, Zhihong and Xie, Yuxin and Liang, Liming and Zou, Yuexian},
  booktitle={Proceedings of the Computer Vision and Pattern Recognition Conference},
  pages={4189--4199},
  year={2025}
}

@inproceedings{kim2025fuzzy,
  title={Fuzzy Contrastive Decoding to Alleviate Object Hallucination in Large Vision-Language Models},
  author={Kim, Jieun and Kim, Jinmyeong and Kim, Yoonji and Cho, Sung-Bae},
  booktitle={Proceedings of the IEEE/CVF International Conference on Computer Vision},
  pages={20572--20581},
  year={2025}
}

@misc{jia2024chatgpt,
      title={Can ChatGPT Detect DeepFakes? A Study of Using Multimodal Large Language Models for Media Forensics}, 
      author={Shan Jia and Reilin Lyu and Kangran Zhao and Yize Chen and Zhiyuan Yan and Yan Ju and Chuanbo Hu and Xin Li and Baoyuan Wu and Siwei Lyu},
      year={2024},
      eprint={2403.14077},
      archivePrefix={arXiv},
}

@article{Shi2025SHIELD, 
    author = {Yichen Shi and Yuhao Gao and Yingxin Lai and Hongyang Wang and Jun Feng and Lei He and Jun Wan and Changsheng Chen and Zitong Yu and Xiaochun Cao},
    title = {SHIELD: an evaluation benchmark for face spoofing and forgery detection with multimodal large language models},
    year = {2025},
    journal = {Visual Intelligence},
}

@article{Zou2025SurveyOA,
  title={Survey on AI-Generated Media Detection: From Non-MLLM to MLLM},
  author={Yueying Zou and Peipei Li and Zekun Li and Huaibo Huang and Xing Cui and Xuannan Liu and Chenghanyu Zhang and Ran He},
  journal={ArXiv},
  year={2025},
}

@inproceedings{liu2024forgeryaware,
  title       = {Forgery-aware Adaptive Transformer for Generalizable Synthetic Image Detection},
  author      = {Liu, Huan and Tan, Zichang and Tan, Chuangchuang and Wei, Yunchao and Wang, Jingdong and Zhao, Yao},
  booktitle   = {Proceedings of the IEEE/CVF Conference on Computer Vision and Pattern Recognition (CVPR)},
  year        = {2024},
}

@inproceedings{yang2025d3,
  title={D3: Scaling Up Deepfake Detection by Learning from Discrepancy},
  author={Yang, Yongqi and Qian, Zhihao and Zhu, Ye and Russakovsky, Olga and Wu, Yu},
  booktitle={Proceedings of the IEEE/CVF Conference on Computer Vision and Pattern Recognition},
  year={2025}
}

@article{wang2025internvl3,
    title={InternVL3. 5: Advancing Open-Source Multimodal Models in Versatility, Reasoning, and Efficiency},
    author={Wang, Weiyun and Gao, Zhangwei and Gu, Lixin and Pu, Hengjun and Cui, Long and Wei, Xingguang and Liu, Zhaoyang and Jing, Linglin and Ye, Shenglong and Shao, Jie and others},
    journal={arXiv preprint arXiv:2508.18265},
    year={2025}
}

@article{bai2025qwen3vl,
      title={Qwen3-VL Technical Report}, 
      author={Shuai Bai and Yuxuan Cai and Ruizhe Chen and Keqin Chen and Xionghui Chen and others},
	  journal={arXiv preprint arXiv:2511.21631},
      year={2025}
}

@misc{qwen3_5,
    title  = {{Qwen3.5}: Towards Native Multimodal Agents},
    author = {{Qwen Team}},
    year   = {2026},
    month  = {February},
    url    = {https://qwen.ai/blog?id=qwen3.5}
}

@misc{mimovl,
      title={MiMo-VL Technical Report}, 
      author={LLM-Core-Team Xiaomi},
      year={2025},
      eprint={2506.03569},
      archivePrefix={arXiv},
      primaryClass={cs.CL},
      url={https://arxiv.org/abs/2506.03569}, 
}

@inproceedings{rohrbach2018object,
  title={Object hallucination in image captioning},
  author={Rohrbach, Anna and Hendricks, Lisa Anne and Burns, Kaylee and Darrell, Trevor and Saenko, Kate},
  booktitle={Proceedings of the 2018 Conference on Empirical Methods in Natural Language Processing},
  pages={4035--4045},
  year={2018}
}

@inproceedings{kaul2024throne,
  title={Throne: An object-based hallucination benchmark for the free-form generations of large vision-language models},
  author={Kaul, Prannay and Li, Zhizhong and Yang, Hao and Dukler, Yonatan and Swaminathan, Ashwin and Taylor, CJ and Soatto, Stefano},
  booktitle={Proceedings of the IEEE/CVF Conference on Computer Vision and Pattern Recognition},
  pages={27228--27238},
  year={2024}
}

@inproceedings{chaubey2026face,
  title={Face-LLaVA: Facial expression and attribute understanding through instruction tuning},
  author={Chaubey, Ashutosh and Guan, Xulang and Soleymani, Mohammad},
  booktitle={2026 IEEE/CVF Winter Conference on Applications of Computer Vision (WACV)},
  pages={2648--2660},
  year={2026},
  organization={IEEE}
}

\clearpage
\section{Supplementary Material}

\setcounter{equation}{0}
\setcounter{figure}{0}
\setcounter{table}{0}
\renewcommand{\theequation}{S\arabic{equation}}
\renewcommand{\thefigure}{S\arabic{figure}}
\renewcommand{\thetable}{S\arabic{table}}

\providecommand{\promptbox}[1]{%
  \vspace{0.35em}
  \noindent
  \fcolorbox{promptborder}{promptblue}{%
    \parbox{\dimexpr\columnwidth-2\fboxsep-2\fboxrule\relax}{%
      \small
      #1
    }%
  }%
  \vspace{0.35em}
}

\makeatletter
\renewcommand\section{\@startsection{section}{1}{\z@}%
  {2.0ex plus .5ex minus .2ex}%
  {1.0ex plus .2ex}%
  {\normalfont\Large\bfseries\raggedright}}
\renewcommand\subsection{\@startsection{subsection}{2}{\z@}%
  {1.5ex plus .4ex minus .2ex}%
  {0.8ex plus .2ex}%
  {\normalfont\large\bfseries\raggedright}}
\makeatother

This supplementary material is organized as follows:

\begin{enumerate}
\renewcommand{\labelenumi}{\Alph{enumi}.}
    \item \textbf{Discussion and Limitations} explores the boundary of our approach.
    \item \textbf{Evaluation Metrics} defines how each reported number is computed, covering the artifact-level Cover, CHAIR, Hal, and $F^{0.5}$ protocol used in
    TriDF and the ACC/F1 protocol used in MMTD-Set.
    \item \textbf{Self-Proposed Regions} experiments with language-level guidance in regional examination instead of the blur-contrastive attention mining in \emph{Look Before You Judge}. 
    \item \textbf{Alternative Region Source} replaces the mined regions with fixed landmark-defined facial components corresponding to the eyes, nose, and mouth, while leaving Stage~3 unchanged.
    \item \textbf{Qualitative Results} shows additional side-by-side comparisons between vanilla inference and our framework. 
    \item \textbf{Implementation Details} specifies the model configuration, attention-extraction procedure, per-stage hyperparameters, and evaluation hardware used across all evaluated backbones.
    \item \textbf{Computational Cost \& Runtime} quantifies the overhead our framework introduces, measured over 100 images by the average number of model invocations, peak GPU memory, and end-to-end inference time per image.
    \item \textbf{Prompt Templates} lists the Stage-3 region-inspection and aggregation prompts.
\end{enumerate}

\section{A. Discussion and Limitations}
\label{sec:limitations}

The blur-contrastive attention prior in our framework should be interpreted as a proposal mechanism for acquiring evidence rather than as a causal attribution of the model's decision. Our framework does not imply that the selected regions are the sole causes of the final prediction. Instead, they serve as probes for identifying the MLLM's evidence-sensitive attention head, which are subsequently leveraged to recognize candidate forensic evidence.

On the other hand, our framework inherits several limitations from its training-free design. First, because candidate evidence regions are proposed through blur-contrastive attention responses, the framework naturally emphasizes artifacts that are sensitive  to local high-frequency perturbations, potentially making it less effective for manipulations characterized by more subtle low-frequency inconsistencies. Future work could explore richer perturbation strategies or adaptive mechanisms to improve coverage across diverse manipulation types. Second, the framework requires white-box access to decoder attention, limiting its applicability to open-weight MLLMs. Extending evidence acquisition to black-box models remains an important direction for future research. Finally, the quality of the acquired evidence depends on the preservation of fine-grained visual details, and performance may degrade under severe compression, downsampling, or coarse visual-token representations. Incorporating multi-scale visual representations or resolution-aware evidence acquisition may further improve robustness in such scenarios. 

Despite these limitations, we believe the proposed evidence acquisition formulation, \emph{look-before-you-judge}, provides a flexible foundation that can accommodate more advanced proposal mechanisms, perturbation strategies, and visual representations as future MLLMs continue to evolve.

\section{B. Evaluation Metrics}
\label{sec:metrics}

We adopt the evaluation protocol of TriDF~\cite{jiang2026tridf} without modification, and report standard binary classification metrics on MMTD-Set~\cite{xu2025fakeshield}. Because the artifact-level metrics are central to our claims about explanation grounding, we restate their definitions here so that our numbers can be interpreted and reproduced without reference to the original paper.

\subsection{Artifact-Level Metrics on TriDF}

\noindent\textbf{Artifact mapping.}
All artifact-level metrics operate on \emph{artifact lists} rather than on raw text. Given a model's free-form response $R^{\mathrm{DF}}$ to a Type-B \textless OEQ\textgreater{} query, an external LLM $\theta$ maps the response onto TriDF's predefined artifact taxonomy $\mathit{Art}=\{\mathit{art}_1,\dots,\mathit{art}_n\}$, producing a mapped artifact list
\begin{equation}
    R^{\mathrm{DF}}_{\mathrm{art}} = \theta\!\left(R^{\mathrm{DF}}\right),
\end{equation}
where each entry records whether the response asserts the presence of the corresponding artifact. The reference list $Y^{\mathrm{DF}}_{\mathrm{art}}$ records the artifacts actually annotated for that sample. This mapping step is what makes the metrics robust to surface wording: two responses describing the same artifact with different phrasing map to the same taxonomy entry. It also means that any claim the mapper cannot align with the taxonomy is discarded before scoring.

\noindent\textbf{Cover} measures the recall of annotated artifacts, i.e.\ how much of the ground-truth evidence the explanation recovers:
\begin{equation}
    \mathit{Cover}(R)
    =
    \frac{\left|R^{\mathrm{DF}}_{\mathrm{art}} \cap Y^{\mathrm{DF}}_{\mathrm{art}}\right|}
         {\left|Y^{\mathrm{DF}}_{\mathrm{art}}\right|}.
\end{equation}

\noindent\textbf{CHAIR}~\cite{rohrbach2018object} measures the proportion of asserted artifacts that are \emph{not} supported by the annotation, and is therefore one minus the precision of the artifact claims:
\begin{equation}
    \mathit{CHAIR}(R)
    =
    1 - \frac{\left|R^{\mathrm{DF}}_{\mathrm{art}} \cap Y^{\mathrm{DF}}_{\mathrm{art}}\right|}
             {\left|R^{\mathrm{DF}}_{\mathrm{art}}\right|}.
\end{equation}

\noindent\textbf{Hal} converts CHAIR into a per-sample indicator, so that its average over a dataset is the percentage of responses containing \emph{at least one} unsupported artifact claim:
\begin{equation}
    \mathit{Hal}(R)
    =
    \begin{cases}
        1 & \text{if } \mathit{CHAIR}(R) \neq 0,\\
        0 & \text{otherwise.}
    \end{cases}
\end{equation}
Hal is thus considerably coarser than CHAIR: a response with a single unsupported claim and a response that is entirely fabricated both receive $\mathit{Hal}=1$. Values close to $1$ should accordingly be read as ``almost every response contains some unsupported claim'' rather than as a severity measure.

\noindent\textbf{$F^{\beta}$-score} combines the two directions into a single indicator. Following THRONE~\cite{kaul2024throne}, TriDF treats precision as twice as important as recall, since false positives in forensic explanations are typically hallucination-driven and more damaging than incomplete coverage. Writing $\mathit{Prec}(R) = 1 - \mathit{CHAIR}(R)$ for the precision of the artifact claims,
\begin{equation}
    F^{\beta}(R)
    =
    \frac{(1+\beta^{2})\,\mathit{Prec}(R)\,\mathit{Cover}(R)}
         {\beta^{2}\,\mathit{Prec}(R) + \mathit{Cover}(R)},
\end{equation}
where $\beta = 0.5$.

\noindent\textbf{Penalty conventions.}
Two cases receive the maximal penalty $\mathit{CHAIR}=1$, and consequently $\mathit{Hal}=1$ and $F^{0.5}=0$: an empty mapped artifact list, $|R^{\mathrm{DF}}_{\mathrm{art}}|=0$, and a manipulated sample classified as authentic. Since the artifact-level metrics require annotated artifacts, they are defined on manipulated samples, whereas accuracy (ACC) is computed over the full real--fake set.

\subsection{Detection Metrics on MMTD-Set}

On MMTD-Set, we report accuracy (ACC) and F1 on the DeepFake and AIGC-Editing subsets, treating “manipulated” as the positive class. We report both metrics because ACC summarizes overall classification performance, whereas F1 focuses on the detection of manipulated samples and better exposes models that are biased toward predicting “authentic.”

\noindent\textbf{Reporting precision.}
Scores for FakeShield~\cite{xu2025fakeshield} are reported to two decimal places in the original publication. We zero-pad them to three decimal places for column alignment with our own measurements and introduce no additional precision.

\section{C. Self-Proposed Regions}
\label{sec:self_regions}

We further examine whether the gains of our framework can be explained merely by multi-stage prompting or textual region decomposition. In this baseline, the MLLM first proposes up to three small localized regions for close inspection. The same region-level inspection and final verification procedure is then applied, but without blur-contrastive region mining or attention reallocation.

As shown in Table~\ref{tab:self-proposed_region}, Self-Proposed Region Inspection improves over vanilla inference, indicating that explicitly decomposing the task into local inspection and final aggregation can provide a useful reasoning structure. However, it does not outperform Prompt-only on ACC, Cover, CHAIR, or $F^{0.5}$, and remains substantially behind our full framework. This suggests that asking the model to propose and inspect its own regions does not reliably identify manipulation-relevant evidence; the selected regions may still be driven by semantic saliency or language priors rather than image-specific forensic cues.

In contrast, our full framework achieves the strongest overall performance across detection, artifact grounding, and hallucination reduction. These results show that effective look-before-you-judge reasoning requires more than textual region decomposition: the model must first discover detail-sensitive, image-specific regions and inspect them under region-specific attention steering.

\begin{table}[t]
\centering
\small
\setlength{\tabcolsep}{2pt}
\resizebox{\columnwidth}{!}{%
\begin{tabular}{lccccc}
\toprule
\multirow{2}{*}{\textbf{Method}}
& \multicolumn{5}{c}{\textbf{TriDF}} \\
\cmidrule(lr){2-6}
& \textit{ACC} $\uparrow$
& \textit{Cover} $\uparrow$
& \textit{CHAIR} $\downarrow$
& \textit{Hal} $\downarrow$
& $F^{0.5}$ $\uparrow$ \\
\midrule
Vanilla
& 0.4176 & 0.0270 & 0.9745 & 1.0000 & 0.0296 \\

Prompt-only
& 0.5113 & 0.1369 & 0.8545 & 0.9927 & 0.1340 \\

Self-Proposed Region Inspection
& 0.5076 & 0.0919 & 0.8870 & 0.9495 & 0.0836 \\

\rowcolor{gray!20}
Ours
& \textbf{0.5458}
& \textbf{0.2239}
& \textbf{0.6407}
& \textbf{0.7875}
& \textbf{0.2564} \\
\bottomrule
\end{tabular}
}
\caption{\textbf{Comparison with Self-Proposed Region Inspection.} Results on TriDF using InternVL-3.5-8B.}
\label{tab:self-proposed_region}
\end{table}

\section{D. Alternative Region Source}
\label{sec:alt_regions}

We further examine whether the gains of our framework can be reproduced by steering the model toward predefined facial components. Such anatomical priors are a well-established source of localized supervision in face forensics: DFD-FCG~\cite{han2025towards} adapts a foundation model for video-based deepfake detection through facial-component guidance, and Face-LLaVA~\cite{chaubey2026face} injects face-region priors into an MLLM through face-region guided cross-attention. Both indicate that directing a model's capacity toward facial parts is beneficial, which makes a facial-component prior the natural non-learned alternative to our mined regions. We therefore construct a facial-region variant that replaces Stages~1--2 with landmark-based eye, nose, and mouth regions, and distributes the recycled attention uniformly within each region. All remaining settings, including the region-inspection prompt, intervention layers, evidence filtering, and final aggregation, are kept identical to the full method.

The facial-region variant substantially improves over vanilla inference across both detection and explanation-grounding metrics. This result shows that decomposing the image into local inspection regions and reallocating attention during region-wise evidence acquisition is itself effective, even when the candidate regions are defined by a simple face-centric prior. In other words, part of the improvement comes from explicitly directing the model to inspect localized visual evidence before making the final judgment.

Nevertheless, the full method consistently performs best across all metrics. Since the downstream inspection and aggregation procedures are unchanged, this remaining gap indicates that the effectiveness of regional attention reallocation depends not only on applying the intervention, but also on where the redistributed attention is directed. Predefined facial components provide useful inspection targets for face-related artifacts, but they may spend the limited region budget on less informative facial areas or fail to cover manipulations located elsewhere in the image. In contrast, our blur-contrastive prior derives image-specific regions from the input itself, allowing the inspection targets to adapt across facial manipulations, edited objects, and regenerated image regions. This adaptability is particularly important for settings such as image editing, where the face may occupy only a small portion of the image and the diagnostically useful evidence may instead appear in non-facial content.

\begin{table}[t]
\centering
\small
\setlength{\tabcolsep}{2pt}

\resizebox{0.9\columnwidth}{!}{%
\begin{tabular}{lccccc}
\toprule
\multirow{2}{*}{\textbf{Method}}
& \multicolumn{5}{c}{\textbf{TriDF}} \\
\cmidrule(lr){2-6}
& \textit{ACC} $\uparrow$
& \textit{Cover} $\uparrow$
& \textit{CHAIR} $\downarrow$
& \textit{Hal} $\downarrow$
& $F^{0.5}$ $\uparrow$ \\
\midrule
Vanilla
& 0.4176 & 0.0270 & 0.9745 & 1.0000 & 0.0296 \\
\quad + Facial Regions
& 0.5176 & 0.2096 & 0.6727 & 0.8014 & 0.2343 \\
\rowcolor{gray!20}
\quad + Ours
& \textbf{0.5458} & \textbf{0.2239} & \textbf{0.6407} & \textbf{0.7875} & \textbf{0.2564} \\
\bottomrule
\end{tabular}
}
\caption{\textbf{Comparison with Alternative Region Selection Method.} Results on TriDF using InternVL-3.5-8B.}
\label{tab:facial_regions}
\end{table}

\section{E. Qualitative Results}
\label{sec:qualitative}

We provide three additional examples of successful cases and one representative failure case. These examples show how region mining and region-wise inspection can correct erroneous global judgments, improve explanation grounding, and calibrate ambiguous local evidence.

\begin{figure*}[p]
    \centering
    \includegraphics[width=0.9\textwidth,height=0.40\textheight,keepaspectratio]
    {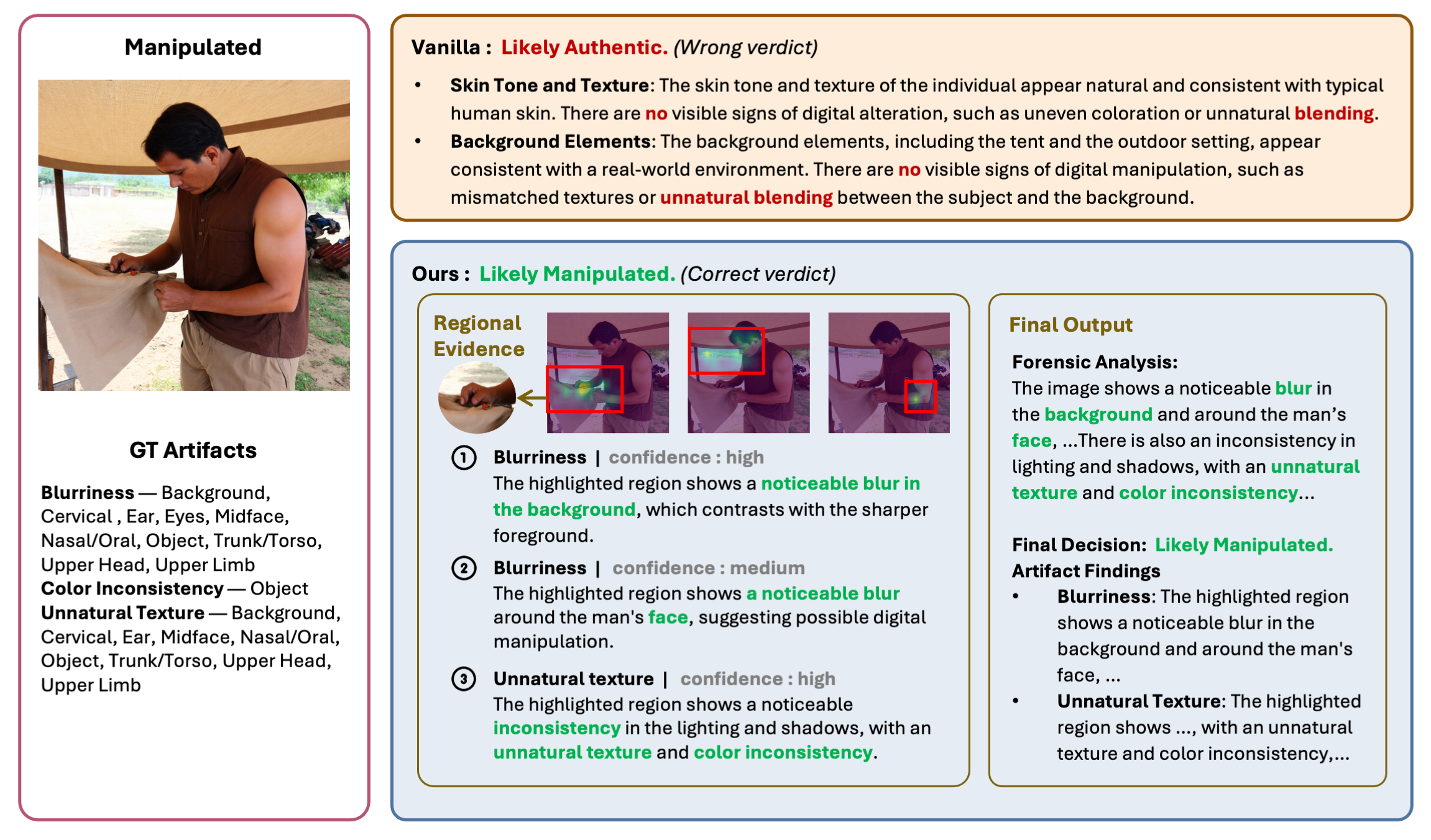}
    \caption{Qualitative result on a manipulated \texttt{image generation} example. Vanilla inference overlooks the localized artifacts and predicts the image as authentic. Our framework identifies blur and texture/color
    inconsistencies in the selected regions and produces the correct
    manipulated verdict.}
    \label{fig:supp_qual_editing}
\end{figure*}

\begin{figure*}[p]
    \centering
    \includegraphics[width=0.9\textwidth,height=0.40\textheight,keepaspectratio]
    {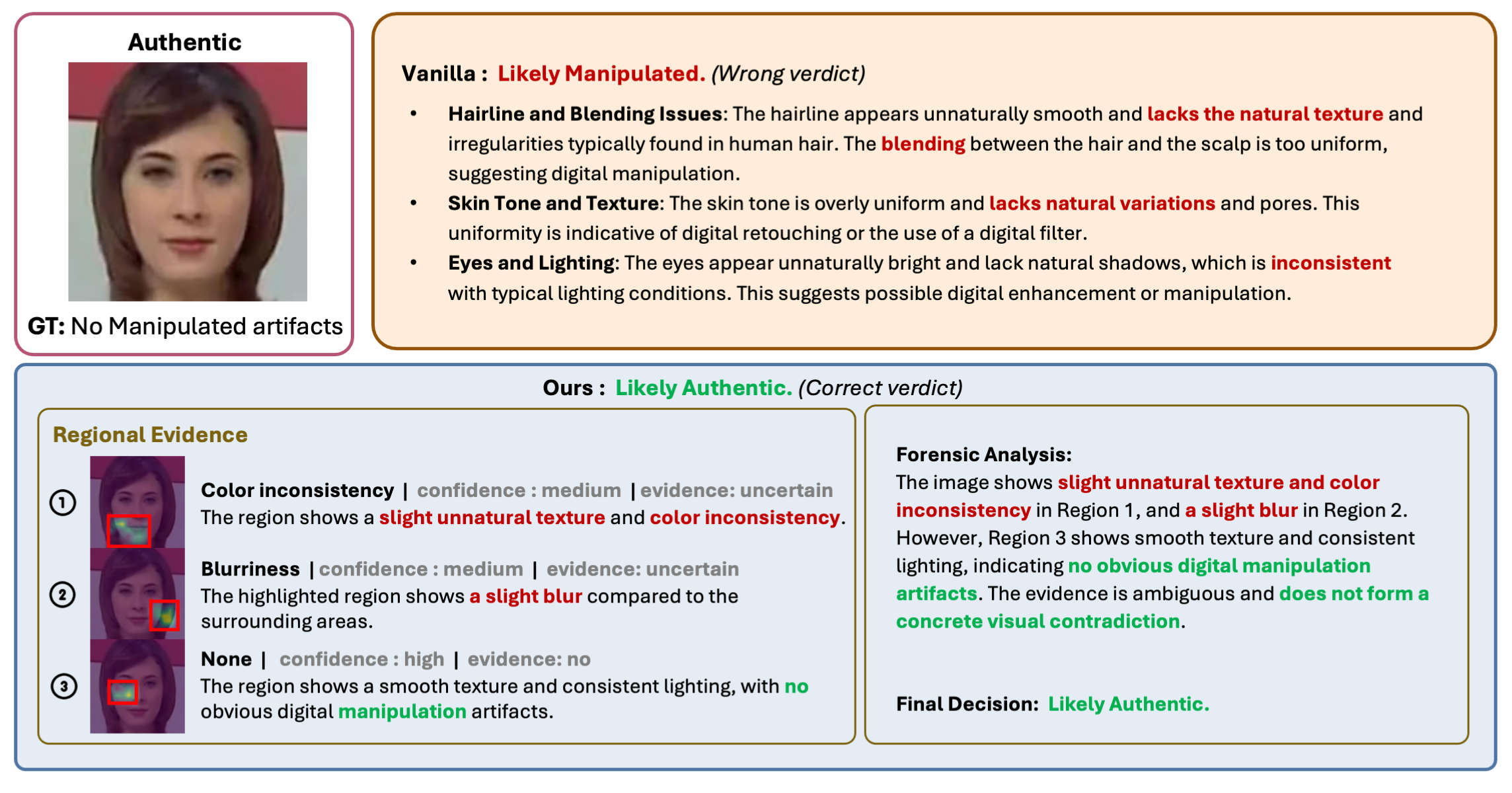}
    \caption{Qualitative result on an authentic face. Vanilla inference hallucinates facial artifacts and produces a false positive. Our framework assigns only weak, uncertain, or negative evidence to the inspected regions and recovers the correct authentic verdict.}
    \label{fig:supp_qual_real}
\end{figure*}

\paragraph{Correcting erroneous global judgments.}
Figures~\ref{fig:supp_qual_editing} and~\ref{fig:supp_qual_real} show complementary cases in which our framework corrects vanilla inference. In the manipulated \texttt{image generation} example, vanilla relies on broad claims of natural appearance and overlooks localized blur and texture inconsistencies. Our framework inspects the discovered regions and integrates the resulting evidence into the correct manipulated verdict.

In the authentic-face example, vanilla instead hallucinates manipulation cues around the hair, skin, and eyes. Although our framework also selects locally ambiguous regions, it describes them with low evidential strength and retains a negative observation. Global aggregation rejects the unsupported cues and produces the correct authentic prediction. Together, these examples show that the method calibrates local evidence rather than simply increasing sensitivity to manipulation.

\begin{figure*}[p]
    \centering
    \includegraphics[width=0.9\textwidth,height=0.40\textheight,keepaspectratio]
    {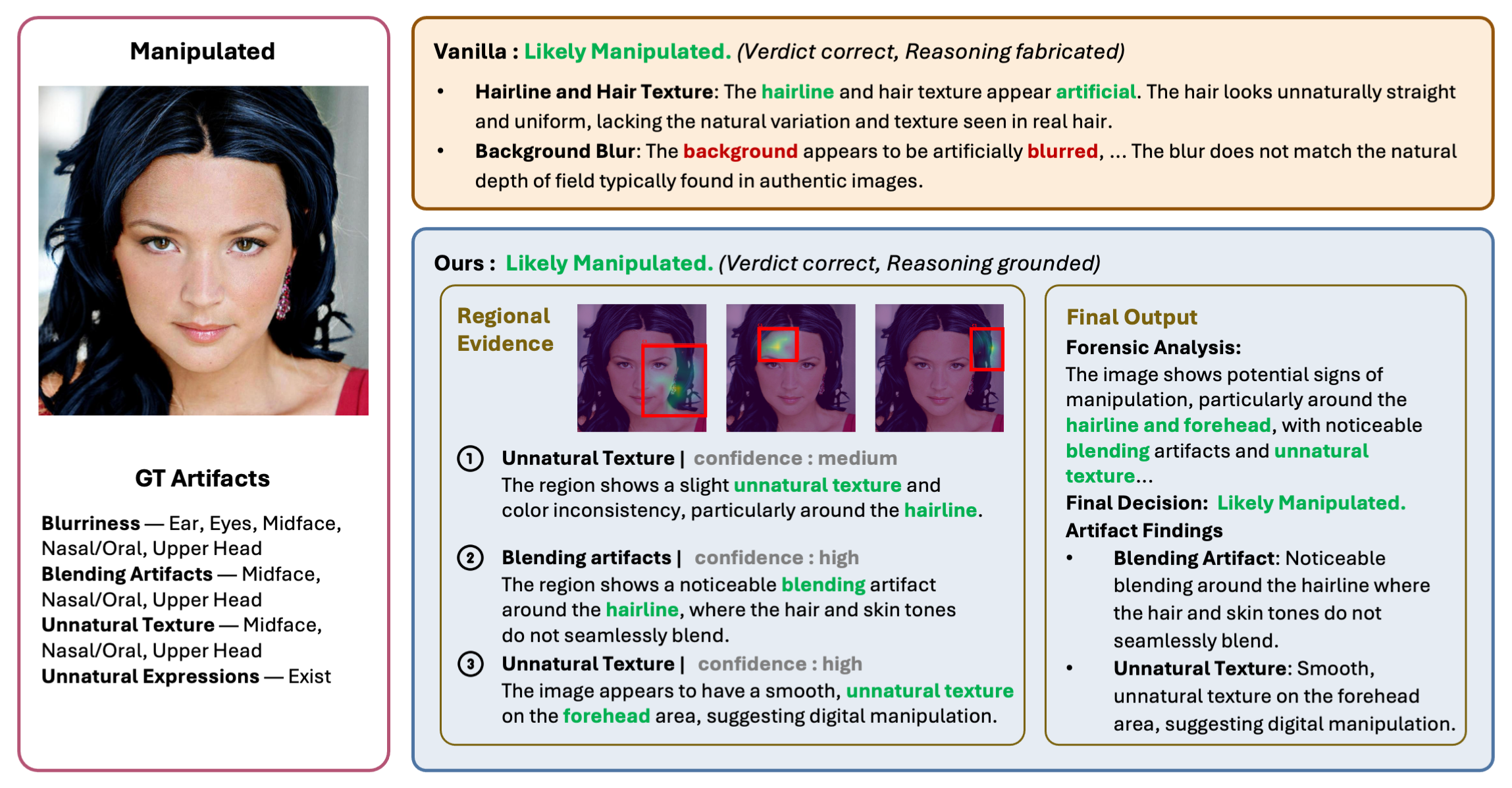}
    \caption{Qualitative comparison when both methods correctly predict manipulation. Vanilla inference provides generic and weakly supported claims, whereas our framework grounds its explanation in blending and texture irregularities around the hairline and forehead.}
    \label{fig:supp_qual_grounding}
\end{figure*}

\paragraph{Improving explanation grounding.}
Figure~\ref{fig:supp_qual_grounding} shows that a correct verdict does not necessarily imply a grounded explanation. Vanilla correctly predicts manipulation but describes generic hair and background anomalies. Our framework instead links its findings to inspected regions around the hairline and forehead, where the observed blending and texture irregularities more closely match the annotated artifacts. The final explanation is therefore more localized and visually traceable.

\begin{figure*}[p]
    \centering
    \includegraphics[width=0.9\textwidth,height=0.40\textheight,keepaspectratio]
    {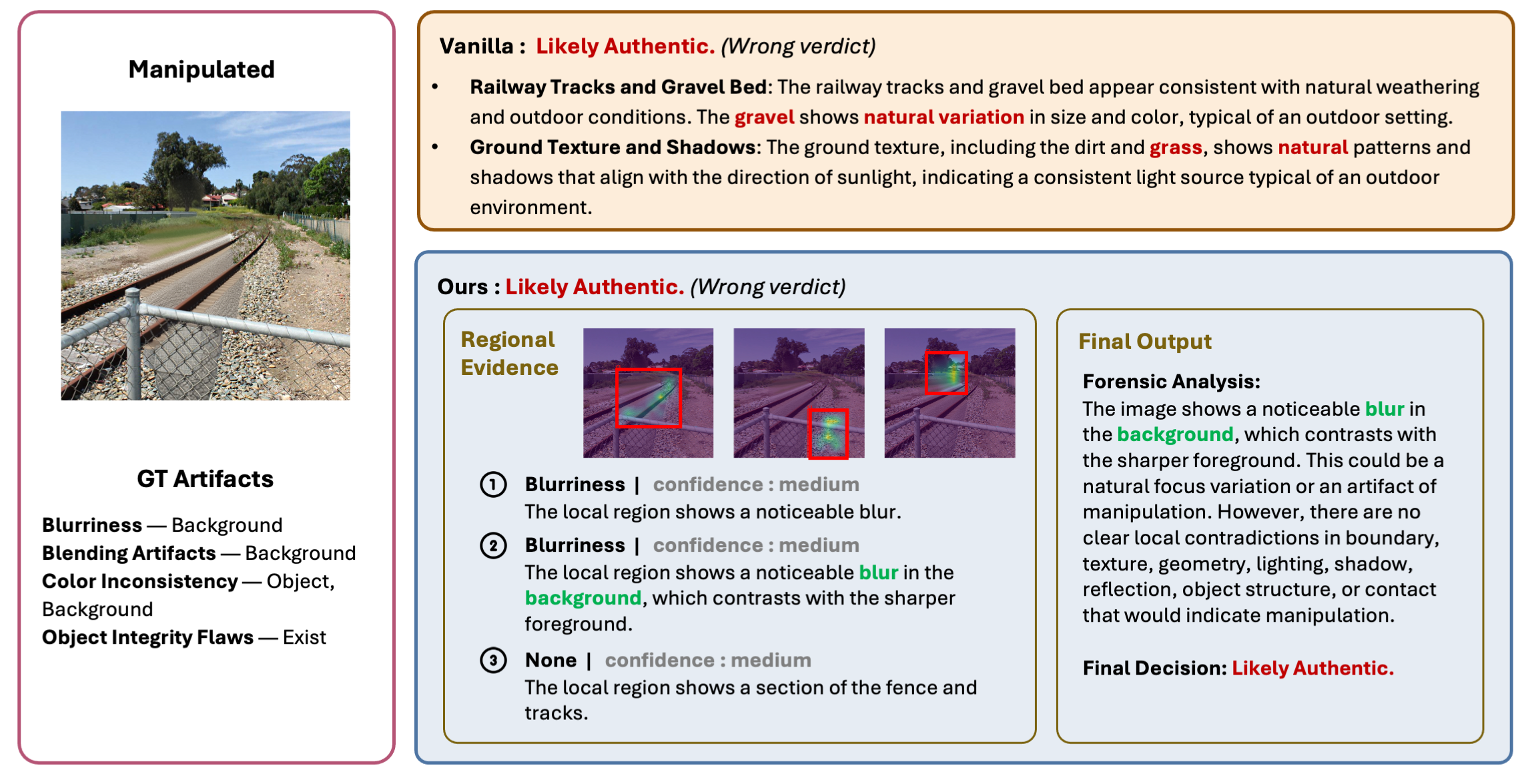}
    \caption{Representative failure case on a manipulated outdoor scene.
    Our framework identifies a relevant background-blur cue, but the cue is
    also compatible with natural focus variation and lacks sufficient
    complementary evidence. Both methods therefore predict the image as
    authentic.}
    \label{fig:supp_qual_failure}
\end{figure*}

\paragraph{Failure case.}
Figure~\ref{fig:supp_qual_failure} presents a manipulated image that both methods classify as authentic. Vanilla supports its decision using broad claims about natural scene appearance. Our framework provides a more localized observation by identifying blur in the railway background, but this cue remains ambiguous because it may also result from natural focus variation. Moreover, the selected regions do not reveal sufficiently strong complementary evidence of blending, color inconsistency, or object-integrity flaws.

This example suggests that the current framework may be less effective when manipulation evidence is spatially diffuse, background-dominated, or naturally explainable in isolation. Nevertheless, its intermediate reasoning remains better calibrated than vanilla inference, as it identifies a relevant local cue without treating it as conclusive evidence.


\section{F. Implementation Details}
\label{sec:implementation}

We instantiate the three-stage framework identically across all evaluated backbones, changing only the quantities that are inherently architecture-dependent: the decoder layer/head counts $L,H$, the visual-token grid $H_v\times W_v$, and the steering layer range $\mathcal L^\star$. All other hyperparameters are shared, apart from two adjustments required by the hybrid attention design of Qwen3.5-9B, which we state explicitly below. This section gives the concrete instantiation for InternVL-3.5-8B, the backbone used for the ablation studies, and describes how the remaining backbones differ.

\noindent\textbf{Model and inference environment.}
InternVL-3.5-8B pairs a $448\times448$ InternViT vision encoder (patch size 14, $0.5\times$ pixel-shuffle downsampling) with a 36-layer, 32-head Qwen3 decoder (8 key--value heads under grouped-query attention). We disable dynamic image tiling (\texttt{max\_num=1}), so every image is encoded as a single tile of $H_v\times W_v=16\times16=256$ visual tokens. All backbones run in \texttt{bfloat16} on a single GPU. Because Stage~1 and Stage~3 both require materialized per-head attention weights, we load the model with FlashAttention2 disabled and request \texttt{output\_attentions=True}; region-wise steering (Stage~3.a) is implemented as a forward-pass patch over the decoder's self-attention module rather than a modification to the generation config. All generation calls -- the Stage~1 fixed continuation, Stage~3.a region inspection, and Stage~3.b final aggregation -- use greedy decoding (\texttt{do\_sample=False}, one beam), so outputs are deterministic given fixed weights and inputs.

\noindent\textbf{Per-stage hyperparameters.}
We instantiate Stage~1 with $L\times H=36\times32$ decoder layers and heads, a Gaussian blur radius of 3.0px defining $I^{\mathrm{blur}}$, and a fixed continuation capped at 64 generated tokens. Head selection retains up to $K_H=20$ heads under fixed diversity constraints (at most 4 heads per layer, minimum grid-distance 2 between selected peaks) and fixed nondegeneracy checks (bounds on border mass, peak mass, effective tokens, and $\Delta^+$ mass); heads failing any check are excluded from $\mathcal H$ regardless of score. Stage~2 operates on the $H_v\times W_v=16\times16$ visual-token grid, retains the top-$p$ fraction of grid cells with $p=0.15$, groups them with 8-connectivity, discards components smaller than 2 grid cells, dilates survivors by one grid step, and keeps up to $K=3$ ranked regions. Stage~3 uses retention coefficient $\rho=0.5$ with sink threshold $1.0$, steers all heads within decoder layers $\mathcal L^\star=[12,31]$, and caps region inspection and final aggregation at 96 and 384 generated tokens, respectively. During evidence filtering (Stage~3.b), an observation is admitted only at medium confidence or higher, and it is dropped as a near-duplicate of an already-kept observation when it shares the same evidence label and artifact type and its normalized text similarity exceeds 0.82. Observations labeled \emph{uncertain} are, by default, retained as weak cues rather than discarded outright, provided they name a concrete artifact type and meet the confidence floor.

\noindent\textbf{Adapting to other backbones.}
The steering layer range $\mathcal L^\star$ and the visual-token grid $H_v\times W_v$ are the two hyperparameters above that are inherently architecture-dependent (Section~A); we identify each once per backbone family by inspecting its \texttt{transformers} implementation and hold it fixed across all of that backbone's evaluations, rather than tuning it per sample. We use $\mathcal L^\star=[12,31]$ (of $L=36$ layers, $H=32$) for InternVL-3.5-8B, $\mathcal L^\star=[19,39]$ (of $L=40$ layers, $H=40$) for InternVL-3.5-14B, $\mathcal L^\star=[18,35]$ (of $L=36$, $H=32$) for Qwen3-VL-8B-Instruct, and the full decoder depth for MiMo-VL-7B. All remaining hyperparameters above are reused unchanged across backbones, with the single exception noted below for Qwen3.5-9B.

\noindent\textbf{Hybrid-attention backbones: Qwen3.5-9B.}
Qwen3.5-9B is architecturally hybrid, and is the one backbone that required an adjustment beyond $\mathcal L^\star$ and the grid. Of its $L=32$ decoder layers, only eight use full self-attention -- layers $\{3,7,11,15,19,23,27,31\}$, with $H=16$ heads each -- while the remaining 24 use linear (GatedDeltaNet) attention and are therefore not steerable under our formulation. We accordingly set $\mathcal L^\star$ to this fixed eight-layer subset and steer all 16 heads of any layer it selects. This leaves only 128 candidate layer--head cells, an order of magnitude fewer than the other backbones, so the shared minimum-peak-distance gate (default 2 grid cells) admits only about 9 of the intended $K_H=20$ heads on this backbone; we therefore lower it to 1 for Qwen3.5-9B alone, which restores selection to 15--17 heads at a modest cost in spatial diversity among the selected heads. Its visual-token grid $H_v\times W_v$ is likewise smaller than the other backbones' and varies slightly across images because preprocessing preserves the input aspect ratio. Finally, because its responses are longer, we raise its Stage~3 region-inspection and final-aggregation budgets to 256 and 512 generated tokens (versus 96 and 384 elsewhere) to avoid truncation. All remaining hyperparameters are unchanged.

\noindent\textbf{Hardware.}
All models are evaluated on a single GPU without model or tensor parallelism. InternVL-3.5-8B, Qwen3-VL-8B-Instruct, Qwen3.5-9B, and MiMo-VL-7B use an NVIDIA RTX 4090 (24\,GB), while InternVL-3.5-14B uses an NVIDIA RTX PRO 6000 Blackwell (96\,GB) to support its larger eager-attention memory footprint. Section~G compares vanilla inference and our framework on the same device.

\section{G. Computational Cost \& Runtime}
\label{sec:computational}

\begin{table}[t]
\centering
\small
\setlength{\tabcolsep}{4pt}
\resizebox{\columnwidth}{!}{%
\begin{tabular}{lcccc}
\toprule
\multirow{2}{*}{\textbf{Method}}
& \multicolumn{4}{c}{\textbf{Inference Efficiency}} \\
\cmidrule(lr){2-5}
& \makecell{\textit{Avg.}\\\textit{Invoc.}$\downarrow$}
& \makecell{\textit{Time}\\\textit{(s/img)}$\downarrow$}
& \makecell{\textit{Rel.}\\\textit{Latency}$\downarrow$}
& \makecell{\textit{Peak GPU}\\\textit{Mem. (GB)}$\downarrow$} \\
\midrule
Vanilla
& 1.000
& 13.480
& 1.000$\times$
& 16.129 \\

\rowcolor{gray!20}
Ours
& 6.970
& 19.951
& 1.480$\times$
& 16.624 \\
\bottomrule
\end{tabular}
}
\caption{\textbf{Inference efficiency comparison.}
inference efficiency on a balanced 100-image subset of TriDF. Both methods use the same InternVL-3.5-8B backbone under identical single-GPU settings. A model invocation denotes either an autoregressive generation or an explicit attention forward evaluation.}
\label{tab:inference_efficiency}
\end{table}

We evaluate inference efficiency on a balanced 100-image subset of TriDF, covering five image manipulation tasks with 10 of each real and fake samples per task. Vanilla inference and our method use the same InternVL-3.5-8B backbone, BF16 precision, $448\times448$ inputs, and single-GPU execution. Model loading, warm-up, and result serialization are excluded, while the same per-sample memory cleanup procedure is retained for both methods. We report the average number of high-level model invocations, inference time per image, relative latency, and peak allocated GPU memory. An invocation denotes either an autoregressive generation or an explicit attention forward evaluation, rather than an individual decoding step.

As shown in Table~\ref{tab:inference_efficiency}, vanilla inference requires one invocation and takes $13.480$ seconds per image, with $16.129$ GB peak memory. Our method averages $6.970$ invocations and takes $19.951$ seconds per image, corresponding to $1.480\times$ relative latency. Peak memory increases by only $0.495$ GB, from $16.129$ GB to $16.624$ GB, or approximately $3.1\%$. The invocation count is slightly below seven because the number of candidate regions is input-dependent. Stage~1 uses three fixed invocations, Stage~2 introduces no model invocation, and Stage~3 uses one inspection call per retained region followed by final aggregation. The total is therefore $4+K$, where $K\leq3$; the observed average corresponds to $2.970$ regions per image.

The increase in invocation count does not produce a proportional increase in runtime because the additional calls are substantially cheaper than a full vanilla generation. Our method distributes a comparable overall generation budget across short intermediate outputs and final aggregation, rather than repeating a full-length response at every stage. In addition, Stage~1 generates the continuation only once and reuses it for the raw and blurred attention evaluations, avoiding an additional autoregressive generation. Stage~2 consists only of lightweight thresholding and connected-component processing, while the regional inspections produce short, structured observations.

The limited memory increase also follows from the sequential design. All stages reuse the same backbone, and the additional attention maps and steering states are transient rather than maintained as parallel model branches. Overall, the proposed framework introduces a moderate $48.0\%$ latency overhead and approximately $3.1\%$ additional peak GPU memory despite requiring nearly seven high-level invocations. This indicates that its computational cost grows substantially more slowly than invocation count while supporting multi-stage region mining and evidence aggregation.

\section{H. Prompt Templates}
\label{sec:prompt}
All evaluated backbones use the same Stage~3 prompting protocol: they inspect only the attention-guided local region, compare it with nearby visual context, and produce structured observations instead of image-level decisions. The retained observations are then verified under a full-image view before generating the final prediction and explanation.

To accommodate differences in instruction following, output-format compliance, and confidence calibration across MLLM backbones, we apply only minor wording adjustments to the shared prompts. These changes do not affect the region proposals, attention steering, evidence schema, filtering rules, or evidence-before-judgment workflow.

\begin{quote}
\promptbox{
\textbf{Prompt S1: Region-Wise Evidence Inspection.}

You are a forensic media authenticity inspector performing a local region-level inspection. Inspect only the attention-guided local region. Do not make the final image-level authentic/manipulated decision at this stage. Examine the strongest visible property within the guided region and compare the local structure with its nearest relevant visual context, such as an adjacent boundary, surface, body part, object, shadow, reflection, or repeated pattern. Focus on concrete and observable visual evidence rather than speculation. Classify the local observation as:

\begin{itemize}
\item \textbf{yes}: a specific localized contradiction consistent with editing or synthesis is clearly visible;
\item \textbf{uncertain}: a visible irregularity exists, but a plausible natural explanation remains;
\item \textbf{no}: the local structure appears visually coherent.
\end{itemize}

Generic blur, smoothness, compression, focus variation, lighting variation, or image quality alone is not sufficient manipulation evidence. The artifact type should describe the concrete visible cue being evaluated rather than an inferred manipulation process. Return exactly:
\begin{itemize}
\item  \textbf{Observation:} $<$one concise sentence describing the local comparison and visible result$>$

\item  \textbf{Manipulation Evidence:} yes / no / uncertain

\item  \textbf{Confidence:} high / medium / low

\item  \textbf{Artifact Type:} $<$short visual artifact or none$>$

\item  \textbf{Natural Explanation:} $<$brief explanation for no or uncertain; otherwise none$>$
\end{itemize}
}
\end{quote}

\begin{quote}
\promptbox{
\textbf{Prompt S2: Global Evidence Aggregation.}

You are a forensic media authenticity inspector. Determine whether the provided image is authentic or manipulated. The region-level observations below are tentative candidate cues rather than verified findings. 

First verify whether each retained cue is visibly supported by the full image and consistent with its surrounding visual context. Give greater weight to specific localized contradictions in boundary, texture, geometry, lighting, shadow, reflection, object structure, occlusion, or physical contact than to generic image properties. 

Weak or ambiguous observations should be treated as lower-confidence cues when a plausible natural explanation remains. Repeated observations describing the same underlying visual effect should be consolidated rather than counted as independent evidence. A mostly natural-looking image does not invalidate a concrete local contradiction, but normal global context may weaken generic or unsupported cues.

Retained region-level observations:

\textbf{{Filtered Region Observations}}

Base the final decision on the specificity, visual support, consistency, and overall strength of the verified evidence.

Return exactly:

\textbf{Forensic Analysis:}

$<$In one or two sentences, summarize which local cues remain visually supported after the full-image verification and how strongly they support the manipulation hypothesis.$>$

\textbf{Final Decision:} Likely Authentic or Likely Manipulated.

\textbf{Artifact Findings:}

\textbf{- Title of artifact:} $<$short visual artifact or none$>$

\textbf{- Reason:} $<$brief technical rationale grounded in visible evidence$>$

}
\end{quote}

\end{document}